\documentclass[10pt,twocolumn,letterpaper]{article}

\usepackage{iccv}
\usepackage{times}
\usepackage{epsfig}
\usepackage{epstopdf}
\usepackage{graphicx}
\usepackage{amsmath}
\usepackage{amssymb}
\usepackage{subfigure}
\usepackage{multirow}
\usepackage{longtable}
\usepackage{rotating}
\usepackage{array}
\usepackage{color}
\usepackage{acronym}
\usepackage[normalem]{ulem}
\usepackage{float}

\usepackage[pagebackref=true,breaklinks=true,letterpaper=true,colorlinks,bookmarks=false]{hyperref}

\newcommand{\comment}[1]{}

\newcommand\remove[1]{}

\newcommand{\x}{\mathbf{x}}
\newcommand{\V}{\mathbf{V}} 
\newcommand{\patchT}{\mathcal{P}_T} 
\newcommand{\patchI}{\mathcal{P}_{\hat{I}}}
\newcommand{\R}{\mathbb{R}} 

\acrodef{NCC}[NCC]{Normalized Cross Correlation}
\acrodef{MI}[MI]{Mutual Information}
\acrodef{SSD}[SSD]{Sum of Squared Differences}

\iccvfinalcopy 

\def\iccvPaperID{2119} 

\ificcvfinal\fi
\begin{document}

\title{\vspace{-6mm} Dense image registration and deformable surface reconstruction \\ in presence of occlusions and minimal texture \thanks{This work was supported in part by the Swiss National Science Foundation and ICT R\&D program of MSIP/IITP [B0101-15-0307]. E-mails: {\it\{firstname.lastname\}@epfl.ch, \{shine0624, cd\_yoo\}@kaist.ac.kr}} \vspace{-4mm}}

\author{
 Dat Tien Ngo$^a$ \hspace{0.3cm}
 Sanghyuk Park$^b$ \hspace{0.3cm}
 Anne Jorstad$^a$  \hspace{0.3cm}
 Alberto Crivellaro$^a$ \hspace{0cm} \\
 \hspace{-0.2cm} Chang Yoo$^b$ 		\hspace{0.7cm}
 Pascal Fua$^a$ \vspace{1mm} \\
 {\normalsize $^a$Computer Vision Laboratory, EPFL, Switzerland} \vspace{-0.0cm}\\
 {\normalsize $^b$School of Electrical Engineering, KAIST, Korea} \\
 \vspace{-9mm}
}

\maketitle

\begin{abstract}
\vspace{-3mm}
Deformable surface tracking from monocular images is well-known to be under-constrained. Occlusions often make the task even more challenging, and can result in failure if the surface is not sufficiently textured.
In this work, we explicitly address the problem of 3D reconstruction of poorly textured, occluded surfaces, proposing a framework based on a template-matching approach that scales dense robust features by a relevancy score.
Our approach is extensively compared to current methods employing both local feature matching and dense template alignment. We test on standard datasets as well as on a new dataset (that will be made publicly available) of a sparsely textured, occluded surface. 
Our framework achieves state-of-the-art results for both well and poorly textured, occluded surfaces.
\end{abstract}


\vspace{-3mm}

\section{Introduction}

Being able to  recover the 3D shape of deformable  surfaces from ordinary images
will make it possible to field reconstruction systems that require only a single
video camera, such as  those that now equip most mobile  devices.  It will also
allow 3D  shape recovery in more  specialized contexts, such as  when performing
endoscopic  surgery or  using a  fast camera  to capture  the deformations  of a
rapidly moving  object. Depth ambiguities  make such monocular shape  recovery 
highly  under-constrained. Moreover,  when the  surface  is  partially
occluded or has minimal  texture, the problem
becomes even more  challenging because there is little or  no useful information
about large parts of it.

Arguably,   these  ambiguities  could   be  resolved  by   using  a
depth-camera, such as the popular Kinect sensor~\cite{Shotton11}.
However,  such depth-cameras  are more  difficult to  fit into  a
cell-phone or an endoscope and have limited range. In  this work,  we focus  on
3D  shape recovery  given a  reference image and a single corresponding 3D
template shape known {\it a priori}.

\begin{figure}[t]
\vspace{-0mm}
\centering
\newcommand{\figureHeight}{2.5cm}
\newcommand{\figureWidth}{3.45cm}
{\subfigure[ ] { \includegraphics[height=\figureHeight,width=\figureWidth]{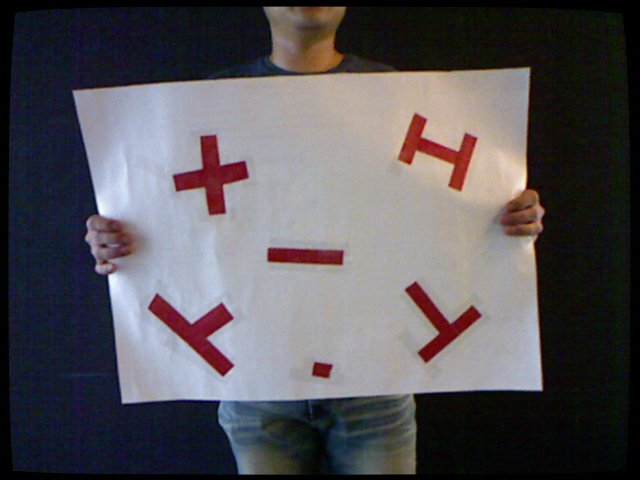}}} \hspace{0.3cm} \vspace{-0.2cm}
{\subfigure[ ] { \includegraphics[height=\figureHeight,width=\figureWidth]{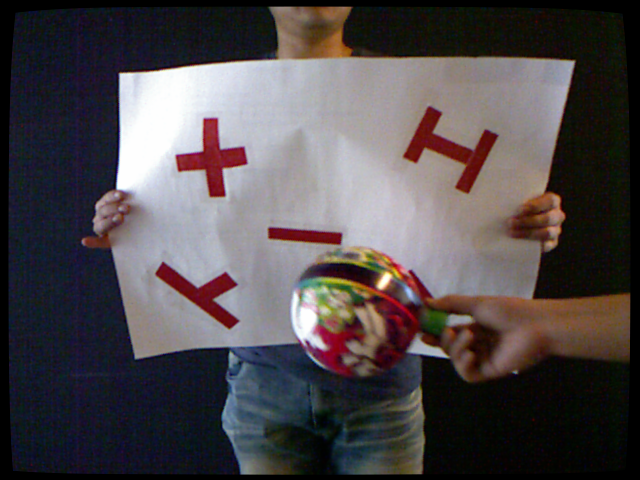} }} \\
{\subfigure[ ] { \includegraphics[height=\figureHeight,width=3.9cm]{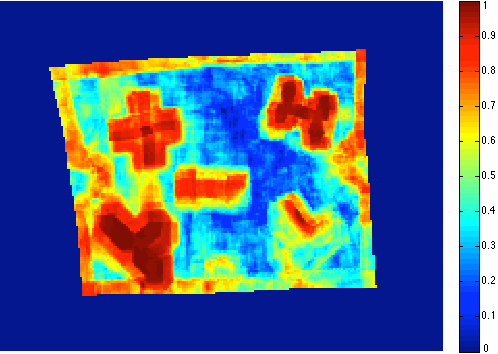} }} \hspace{-0.2cm} \vspace{-0.1cm}
{\subfigure[ ] { \includegraphics[height=\figureHeight,width=\figureWidth]{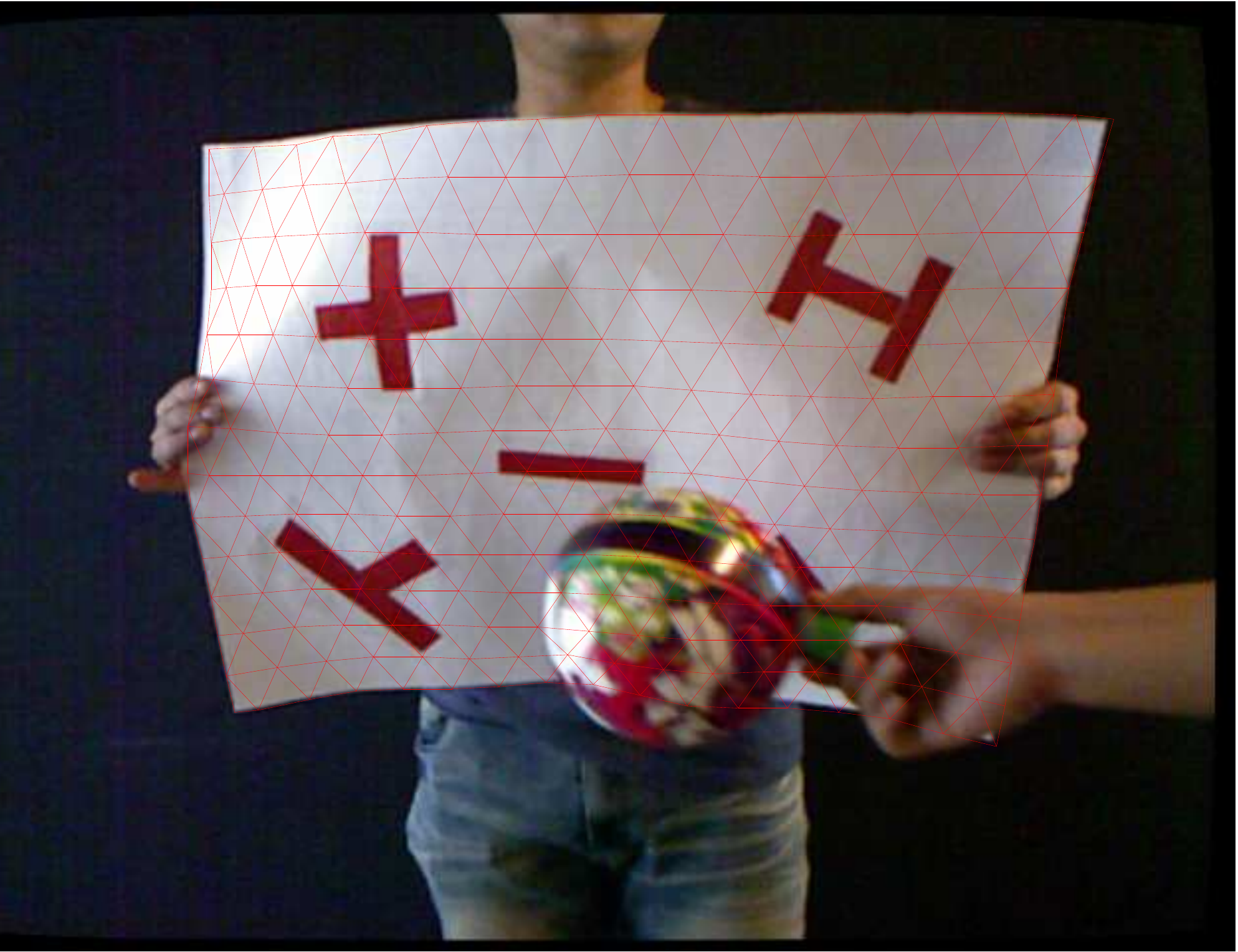} }}
\caption{Tracking a sparsely textured surface in the presence of occlusion:
(a) template image, (b) input image, (c) relevancy score, (d) surface tracking
result with proposed framework. All figures in this paper are best viewed in
color.}
\label{fig_relevancy_scores_low_texture}
\vspace{-5mm}
\end{figure}

When  the surface  is well-textured, correspondence-based  methods have  proved
effective    at   solving    this    problem, even in the presence of occlusions
\cite{Bartoli12b,Bronte14,Brunet14,Chhatkuli14,Perriollat11,Pizarro12,Vicente12}.
In contrast, when  the surface lacks  texture, dense pixel-level  template
matching should  be used  instead. Unfortunately, many  methods such
as~\cite{Malti11,Salzmann08a} either are hampered by a  narrow basin of
attraction, which means they  must  be initialized  from  interest  points
correspondences, or require supervised learning to  enhance robustness.  Using
Mutual Information has often been claimed~\cite{Dame12,Dowson06,Panin08,Viola97}
to  be effective at handling these difficulties but our experiments do not bear
this out. Instead, we advocate template matching over robust dense features that
relies on a pixel-wise relevancy score pre-computed for each frame, as shown in
Fig.~\ref{fig_relevancy_scores_low_texture}. Our approach can handle occlusions
and lack of texture simultaneously. Moreover, no training step is required as in
\cite{Salzmann08a}, which we consider to be an advantage because this obligates
either collecting training data or having sufficient knowledge of the surface
properties, neither of which may be forthcoming.


Our  main  contribution  is  therefore  a  robust  framework  for image
registration and  monocular  3D reconstruction of deformable surfaces in  the
presence of occlusions and minimal texture. A main ingredient is the pixel-wise
relevancy score we use to achieve the robustness.  We will  make the code
publicly available, and  release the dataset we  used to  validate  our 
approach, which  contains  challenging sequences  of sparsely-textured deforming
surfaces and the corresponding ground truth.

\section{Related Work}  
\label{sec_related_work}

%

The main approaches for deformable surface reconstruction either require 2D tracking throughout a batch of images or a video sequence \cite{Akhter11,Garg13,Russell12a} \remove{(a problem referred to as \textit{Non-Rigid Structure from Motion}), }or they assume a  reference template and corresponding 3D shape is known. In this work, we focus on the second approach, which we refer to as \textit{template-based reconstruction}.

The most successful  current approaches generally rely on  finding feature point
correspondences
\cite{Ostlund12,Bartoli12b,Bronte14,Chhatkuli14,Perriollat11,Pizarro12,
  Vicente12}, because they are robust to occlusions.  Unfortunately, as shown by
our experimental  results, these methods tend  to break down when  attempting to
reconstruct sparsely  or repetitively  textured surfaces, since  they rely  on a
fairly high number of correct matches.

Pixel-based techniques are able to overcome some limitations of local feature matching, since they reconstruct surfaces based on a global, dense comparison of images. On the other hand, some precautions must be taken to handle occlusions, lighting changes, and noise. \cite{Malti11} estimates a visibility mask on the reconstructed surface, but unlike us, only textured surfaces and self-occlusions are handled. \cite{GayBellile10} registers images of deformable surfaces in 2D and shrinks the image warps in self-occluded areas. \cite{Bartoli12b} proves that an analytical solution to the 3D surface shape can be derived from this 2D warp. However, the surface shape in self-occluded areas is undefined.
\cite{Collins14b} registers local image patches of feature point correspondences to estimate their depths, and geometric constraints are imposed to classify incorrect feature point correspondences.  In contrast to these local depth estimations, our method reconstructs surfaces globally in order to be more robust to noise and outliers.

Other recent approaches employ supervised learning for enhancing performance \cite{Salzmann13,Varol12a}. In \cite{Salzmann08a} strong results are achieved with poorly textured surfaces and occlusion by employing trained local deformation models, a dense template matching framework using \ac{NCC} \cite{Scandaroli12} and contour detection.  Our proposed framework manages to achieve similar performance without requiring any supervised learning step, while the use of robust, gradient-based dense descriptors recently proposed in \cite{Crivellaro14} avoids the need to explicitly detect contours.

Other techniques employed for dealing with occlusions and noise, such as \ac{MI} \cite{Damen12,Dowson06,Panin08,Viola97} and robust M-estimators \cite{Arya07} are studied explicitly in our context, and found to be successful only up to a point.

Our method is similar to that of~\cite{Pilet07}, where a template matching approach is employed and a visibility mask is computed for the pixels lying on the surface, but in this work a very good initialization from a feature point-based method is required in order for its EM algorithm to converge.
In addition to the geometrical degrees of freedom of the surface, local illumination parameters are explicitly estimated in \cite{Hager98,Silveira07}. This requires a reduced deformation model for the surface to keep the size of the problem reasonable. 

In the proposed framework, we achieve good performance without the need to explicitly estimate any illumination model, so that an accurate  geometric model for the surface can be employed. Furthermore, rather than estimating a simple visibility mask as is often done in many domains such as stereo vision \cite{Strecha06}, face recognition \cite{Zhou09}, or pedestrian detection \cite{Wang09c}, we employ a real-valued pixel-wise relevancy score, penalizing at the same time pixels with unreliable information originating both from occluded and low-textured regions. Our method has a much wider basin of convergence and we can track both well and poorly textured surfaces without requiring initialization by a feature point-based method.

\section{Proposed Framework}  \label{sec_framework}

In this work, we demonstrate that a carefully designed dense template matching framework can lead to state-of-the-art results in monocular reconstruction of deformable surfaces\comment{, without the need for any statistical learning step}. In this section we describe our framework, based on a recently introduced gradient-based pixel descriptors \cite{Crivellaro14} for robust template matching and the computation of a relevancy score for outlier rejection.

%

\subsection{Template Matching}


We assume we are given both a template image $T$ and the rest shape of the corresponding deformable surface, which is a triangular mesh defined by a vector of $N_v$ vertex coordinates in 3D, $\mathbf{V}_T\in\R^{N_v \times 3}$.  
To recover the shape of the deformed surface in an input image {\it I}, the vertex coordinates $\mathbf{V}_T$ of the 3D reference shape must be adjusted so that their projection onto the image plane aligns with {\it I}.

We assume the internal parameters of the camera are known and, without loss of generality, that the world reference system coincides with the one of the camera.\comment{All RGB color images are converted to grayscale before feature descriptors are extracted.}
In order to register each input image, a pixel-wise correspondence is sought between the template and the input image.  Each pixel $\mathbf{x}\in\R^2$ on the template corresponds to a point $\mathbf{p}\in\R^3$ on the 3D surface. This 3D point is represented by fixed barycentric coordinates which are computed by backprojecting the image location $\mathbf{x}$ onto the 3D reference shape.

\begin{figure}
  \vspace{-0mm}
  \centering
  \includegraphics[height=3.4cm]{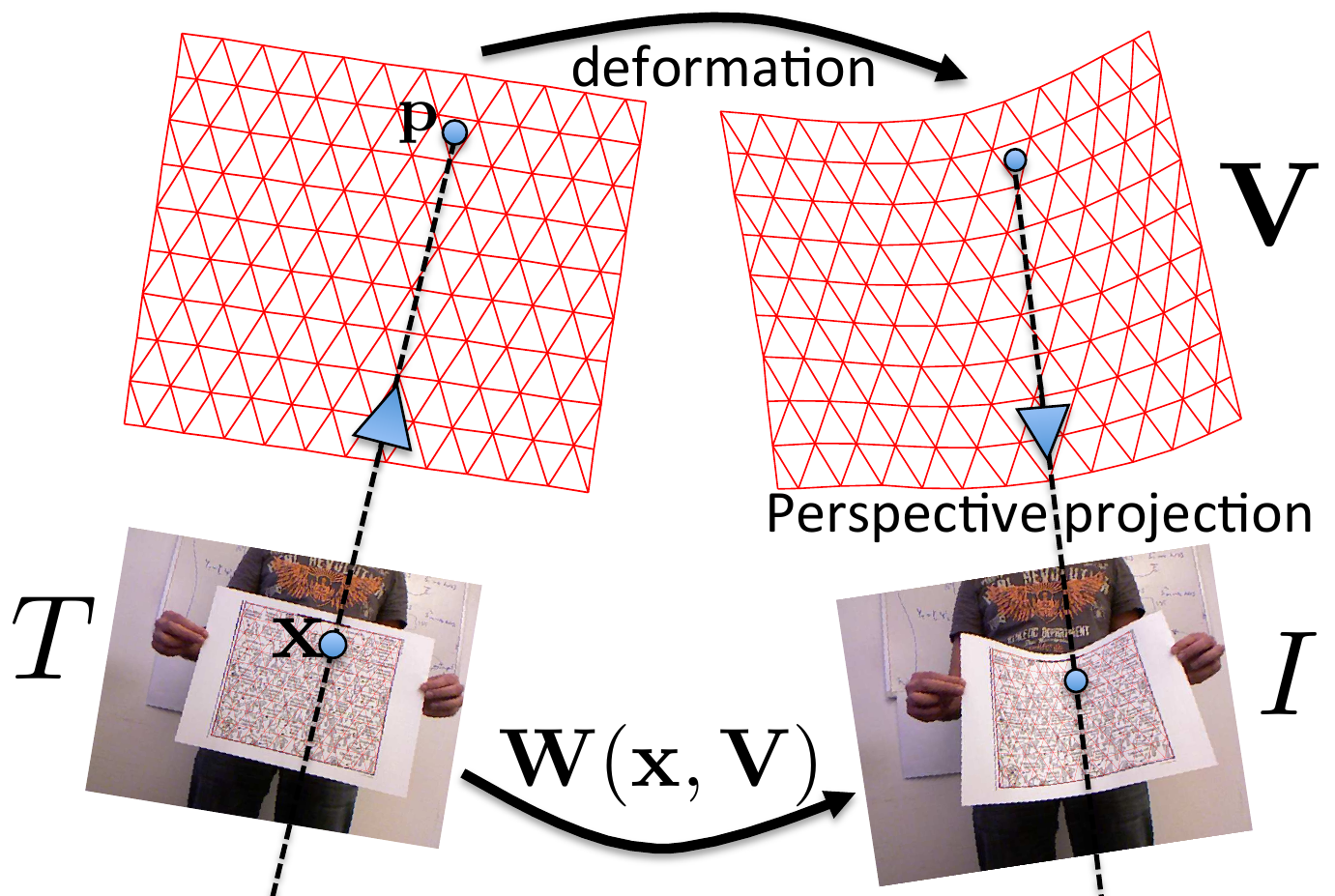}
  \caption{An image warping function maps a pixel from the template image onto the deforming surface in the input image.}
  \label{fig:warp_function}
  \vspace{-3mm}
\end{figure}

The camera projection defines an image warping function $\mathbf{W}: \R^2\times \R^{3\times N_v}\rightarrow \R^2$  which sends pixel $\x$ to a new image location based on the current surface mesh $\mathbf{V}$ as illustrated in Fig.~\ref{fig:warp_function}.
The optimal warping function should minimize the difference between $T(\mathbf{x})$ and $I(\mathbf{W(x;V)})$, according to some measurement of pixel similarity.  Traditionally, image intensity has been used, but more robust pixel feature descriptors $\phi_I(\mathbf{x})$ will lead to more meaningful comparisons, as discussed in Section \ref{features}.




The image energy cost function is a comparison between $\phi_T(\mathbf{x})$ and $\phi_I(\mathbf{W(x;V)})$ at every image point $\mathbf{x}$ defining the quality of their alignment
\begin{equation}
E_{\text{image}}(\V) = \sum_{\x} d\big(\phi_T(\mathbf{x}),
  \phi_I(\mathbf{W(x;V)}) \big) .
  \label{eq:minImageEnergy}
\end{equation}
There are many possible choices for the function $d$ comparing the descriptor vectors, such as \ac{SSD}, \ac{NCC}, \ac{MI}, and others.  We will discuss more in detail about the choice of $d$ in Section \ref{sec_sim_fcn}.


Since monocular 3D surface reconstruction is an under-constrained problem and there are multiple 3D shapes having the same reprojection on the image plane, minimizing the image energy in Eq.~\eqref{eq:minImageEnergy} alone is ill-posed.  Additional constraints must be added, such as isometric deformation constraints enforcing that the surface should not stretch or shrink.  A change in the length between vertex $\mathbf{v}_i$ and vertex $\mathbf{v}_j$ as compared to the template rest length $l_{ij}$ from $\mathbf{V}_T$ is penalized as
\begin{equation}
  E_{\textrm{length}}(\V) = \sum_{i,j}(\left \| \mathbf{v}_i - \mathbf{v}_j \right \| - l_{ij})^2 .
  \label{eq:lengthEnergy}
\end{equation}

To encourage physically plausible deformations, the Laplacian mesh smoothing proposed in \cite{Ostlund12} is used. This rotation-invariant curvature-preserving regularization term based on Laplacian smoothing matrix $\mathbf{A}$ penalizes non-rigid deformations away from the reference shape, based on the preservation of affine combinations of neighboring  vertices.
\begin{equation}
  E_{\text{smooth}}(\V) = \left \| \mathbf{AV} \right \|^2 .
  \label{eq:smoothEnergy}
\end{equation}

To reconstruct the surface, we therefore seek the mesh configuration $\V$ that minimizes the following total energy:
\begin{equation}
    \underset{\mathbf{V}}{\arg\min} \ \  E_{\text{image}}(\V) + \lambda_{\text{L}} E_{\text{length}}(\V) + \lambda_{\text{S}} E_{\text{smooth}}(\V) ,
  \label{eq:totalEnergy}
\end{equation}
for relative weighting parameters $\lambda_{\text{L}}$ and $\lambda_{\text{S}}$.

\subsection{Robust Optimization}  
\label{sec_features}

\subsubsection{Optimization Scheme}
To make the optimization more robust to noise and wide pose changes, we employ a
multi-scale approach, iteratively minimizing  $E^\sigma = 
E_{\text{image}}^\sigma  + \lambda_{\text{L}} E_{\text{length}} +
\lambda_{\text{S}} E_{\text{smooth}}$ for decreasing values of a scale parameter
$\sigma$, with:
\begin{equation}
E_{\text{image}}^\sigma = \sum_{\x} d\left(G^\sigma * \phi_T(\mathbf{x}),
  G^\sigma * \phi_I(\mathbf{W(x;V)}) \right),
  \label{eq:minImageEnergySmooth}
\end{equation}
where $G^\sigma$ is a  low-pass Gaussian filter of variance $\sigma^2$.
In our experiments we solve the alignment at three scales, using the final
result of each coarser scale to initialize the next set of iterations, and
initializing the coarsest scale with the final position found for the previous
frame. The first frame of each image sequence is taken as the template, and we
employ a standard Gauss-Newton algorithm for minimization. \remove{Our
method uses the forward additive registration algorithm, which requires the
computation of the Hessian and Jacobian matrices at every iteration. However, 
we could use the inverse compositional framework algorithm to speed things 
up by pre-computing them only once as in \cite{Brunet11}.}

\subsubsection{Feature Selection}
\label{features}
The image information compared in Eq.~\eqref{eq:minImageEnergy} comes from
pixel-based image features. Previous approaches
\cite{Malti11,Pilet07,Salzmann08a} employ image intensity as a local descriptor,
$\phi_I(\x) = I(\x)$.  More robust results can be obtained with other features,
such as the lighting-insensitive image gradient direction (GD) \cite{Gopalan10},
where $\phi_I(\x) = \tan ^{-1} \frac{I_y(\x)}{I_x(\x)}$ with $\mod 2\pi$
differencing.  Based on its strong previous performance we also consider the
Gradient Based Descriptor Fields (GBDF) recently proposed in
\cite{Crivellaro14}:
\begin{equation}
 \phi_I(\x) = \big[ [ \frac{\partial I }{\partial x}(\x)]^+,\ [\frac{\partial I} {\partial x}(\x)]^-,\ [\frac{\partial I }{\partial y}(\x)]^+,\ [\frac{\partial I} {\partial y}(\x)]^-  \big]^\top \> ,  \label{MC_eqn}
\end{equation}
where the $[\cdot]^+$ and $[\cdot]^-$ operations respectively keep the positive
and negative values of a real-valued signal. These descriptors are robust under
light changes, and remain discriminative after the Gaussian smoothing employed
in Eq.~\eqref{eq:minImageEnergySmooth}; however, as originally proposed in
\cite{Crivellaro14}, they are not rotation invariant. To achieve in-plane
rotation invariance, in our final framework we employ a modified version of
GBDF.  In order to compare pixel descriptors in the same, unrotated coordinate
system, the reconstruction of the previous frame is used to establish a local
coordinate system for each mesh facet.  Each pixel descriptor on the template is
then rotated in accordance with its corresponding mesh facet, to be directly
comparable to the points in the input image.  We show in Section
\ref{sec:experiments} that this modification indeed increases registration
accuracy by being able to successfully track a rotating deformable surface.


\subsubsection{Similarity Function Selection}  \label{sec_sim_fcn}
Choosing the correct comparison function $d$ for Eq.~\eqref{eq:minImageEnergy} also significantly affects the robustness of the tracking. 
Common choices include the \ac{SSD} of the descriptors, and the \ac{NCC} of image intensities \cite{Lewis95}, which is invariant under affine changes in lighting.

\paragraph{Mutual Information:}
\ac{MI} \cite{Viola97} is a similarity function that measures the amount of information shared between two variables, and it is known to be robust to outliers such as noise and illumination changes~\cite{Dame12}. 
It has been repeatedly claimed to be robust to occlusion, for example in~\cite{Dame12,Dowson06,Panin08,Viola97}. Where occlusions occur, the shared information between occluded pixels and the template image is low or none, and its variation does not cause significant change in the image entropy; therefore, the MI obtains an accurate maximum value at the position of the correct alignment, in spite of the occlusion.

However, an MI-based cost function is limited in application.  MI generally provides a non-convex energy function with a very strong response at the optimum, but a very narrow basin of convergence, as shown in Fig.~\ref{fig:exp_align}.  This makes it unsuited for direct numerical optimization, while smoothing leads to a significantly degraded energy function. Numerical experiments reported in Section \ref{sec:experiments} show that, in our context, MI leaves room for improvement.

\paragraph{Robust Statistics:}

M-estimators are a popular method for handling outliers in a template matching framework. Let $e_{i} = \phi_I(\mathbf{W}(\x_i; \V)) - \phi_T(\x_i)$ be the residual at pixel $\x_i$; then instead of minimizing the sum of squared residuals $\sum_{i} e_{i}^2$, a modified loss function $\rho$ of the residuals is considered, instead minimizing $\sum_{i} \rho (e_{i})$, in order to reduce the influence of outliers.

In Section \ref{sec:experiments}, tests are performed using two of the most commonly employed M-estimators, the Huber~\cite{Huber81} and the Tukey~\cite{Hoaglin83} estimators.
In our context, M-estimators show moderate efficacy, likely because part of the useful information is rejected as outliers. This problem becomes particularly significant when dealing with low-textured surfaces, where the amount of information available for alignment is low.

\subsection{Handling Occlusions with a Relevancy Score}  
\label{sec:newMethod}

\begin{figure}
\vspace{-0.2cm}
\centering 
\newcommand{\figureHeight}{1.3cm} 
{\subfigure[]
{\includegraphics[height=\figureHeight]{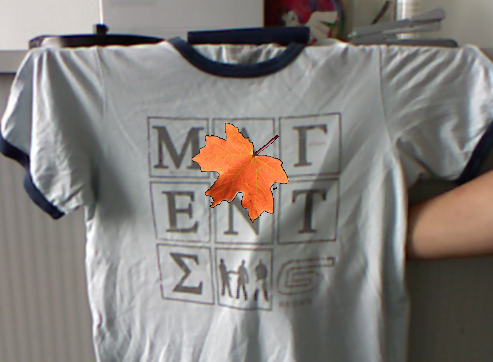}
\hspace{-.1cm}}}
{\subfigure[]
{\includegraphics[height=\figureHeight]{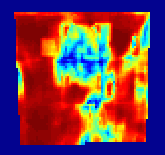}
\hspace{-.1cm}}}
{\subfigure[]
{\includegraphics[height=\figureHeight]{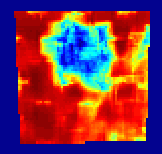}
\hspace{-.1cm}}}
{\subfigure[]
{\includegraphics[height=\figureHeight]{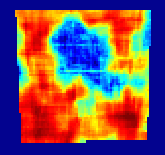}
\hspace{-.1cm}}}
{\subfigure[]
{\includegraphics[height=\figureHeight]{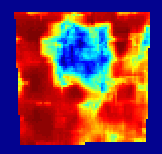}
 \includegraphics[height=\figureHeight]{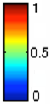}
\hspace{-.1cm}}} \\
\vspace{-2mm}
\newcommand{\figureHeightb}{1.12cm} 
{\subfigure[]
{\includegraphics[height=\figureHeightb]{fig/white_paper/frame_226}
\hspace{-.1cm}}}
{\subfigure[]
{\includegraphics[height=\figureHeightb]{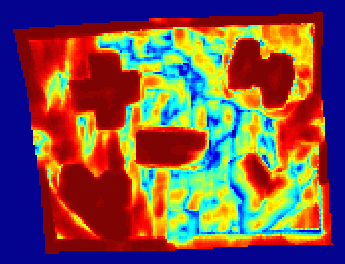}
\hspace{-.1cm}}}
{\subfigure[]
{\includegraphics[height=\figureHeightb]{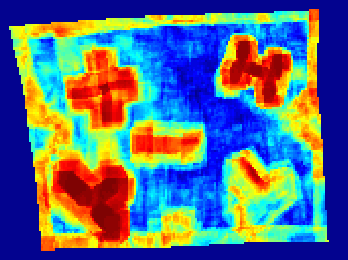}
\hspace{-.1cm}}}
{\subfigure[]
{\includegraphics[height=\figureHeightb]{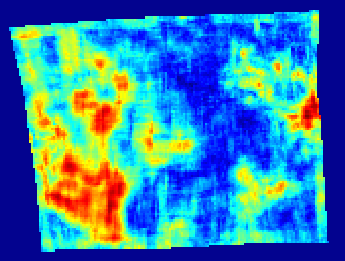}
\hspace{-.1cm}}}
{\subfigure[]
{\includegraphics[height=\figureHeightb]{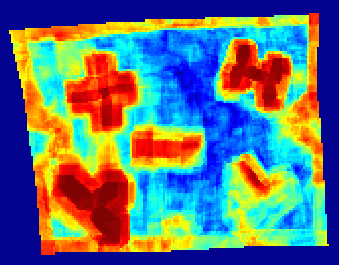}
 \includegraphics[height=\figureHeightb]{fig/colorbar}
\hspace{-.1cm}}}
\caption{The relevancy score results using various methods on the (a) cloth dataset and (f) sparsely textured paper dataset using (b)(g) Intensity (c)(h) GBDF (d)(i) Gradient direction (e)(j) GBDF+Intensity. GBDF gives a better relevancy map than intensity and gradient direction on the first dataset while intensity is better than GBDF and gradient direction on the second. We therefore combine both intensity and GBDF in our proposed relevancy score.}
\label{fig:exp_OCS}
\vspace{-0.4cm}
\end{figure}

Our experiments suggest that selecting a robust similarity function is not enough to deal with the occlusions and image variability encountered when attempting to track a deforming surface in real-world imagery.

Inspired by the effectiveness of the occlusion masks developed in works such as \cite{Strecha06,Wang09c,Zhou09}, we derive a more robust method to handle occlusions by pre-computing a relevancy score for each pixel of the current frame, which is then used to weight the pixels during the alignment. Since we would like to handle occlusions and sparsely textured surfaces together, rather than designing a binary occlusion prediction mask, we develop a continuous-valued score that will raise or lower the importance of pixels depending on their relevancy. This pre-processing step can greatly improve the quality of the image information handed to the cost function in Eq.~\eqref{eq:totalEnergy}.  

Given the estimated configuration $\mathbf{V}^*_{t-1}$ of the deformable surface from the previous frame, a thin-plate spline-based warping function \cite{Bookstein89} is used to un-warp image $I_{t}$ to closely align with the template $T$.  A relevancy score is then computed between each pixel $\mathbf{x}$ on the synthetically back-warped image $\hat{I}_t$ and the same pixel on the template $T$, with a sliding-window approach.

It has been verified repeatedly in the literature that NCC is a reliable choice for measuring patch-based image similarity, and so we compute the NCC over the images as an efficient prediction of relevancy.  
In one such approach \cite{Klein07}, local image patches are affinely warped based on the predicted camera pose, and sliding NCC windows are then used to look for correspondences of map points in the input image.  Our approach is somewhat different, as we use sliding NCC to measure the relevancy of template pixels on the input image.
We average the NCC of both the image intensity and the GBDF features, as it was found that both descriptors provide relevant and often complementary information at this predictor stage, (see Fig.~\ref{fig:exp_OCS} for a qualitative comparison).

The sliding relevancy score is computed as the maximum NCC value over a range of patches near image location $\x$:
\begin{equation}
\omega(\x) = \max_{\mathbf{\delta}} \ \text{NCC}(\patchT(\x), \patchI(\x + \mathbf{\delta})),
\end{equation}
where $\patchT(\x)$ and $\patchI(\x)$ are patches of size $26 \times 26$ centered at $\x$, $\delta = [ \delta_x, \delta_y ]^T $, and $\delta_x, \delta_y$ vary over $[-30,30]$ in all our experiments.  Allowing the patch to be compared to nearby patches accounts for some of the variability between the surface position $\mathbf{V}^*_{t-1}$ and the desired position $\mathbf{V}^*_{t}$ to be recovered.

\begin{figure}
\centering 
\newcommand{\figureWidth}{1.6cm} 
\includegraphics[width=\figureWidth]{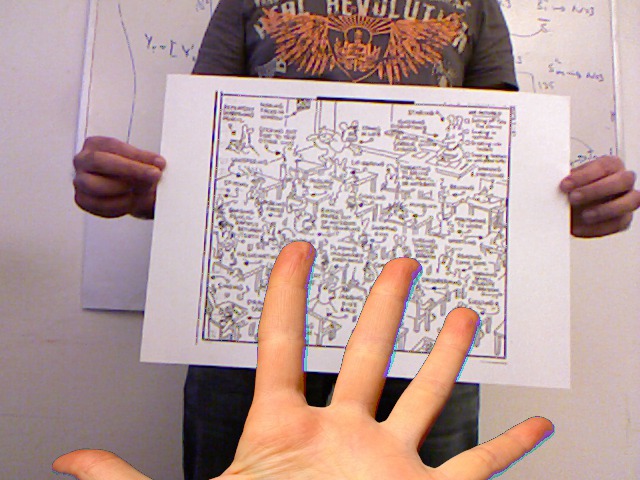}
\includegraphics[width=\figureWidth]{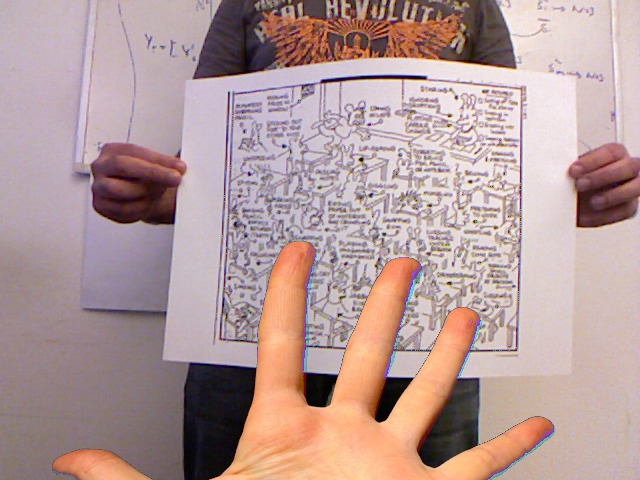}
\includegraphics[width=\figureWidth]{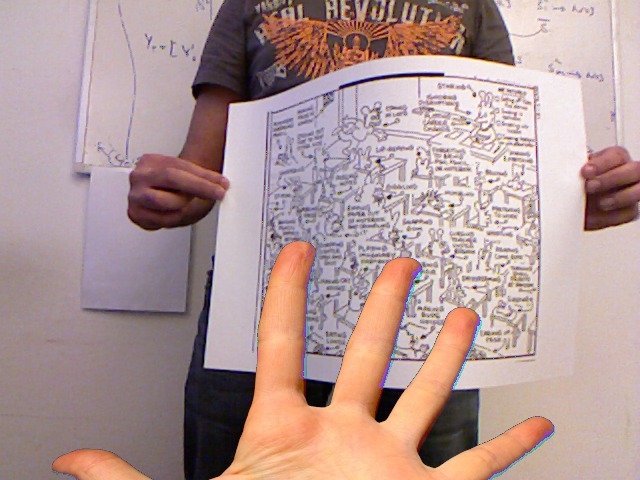}
\includegraphics[width=\figureWidth]{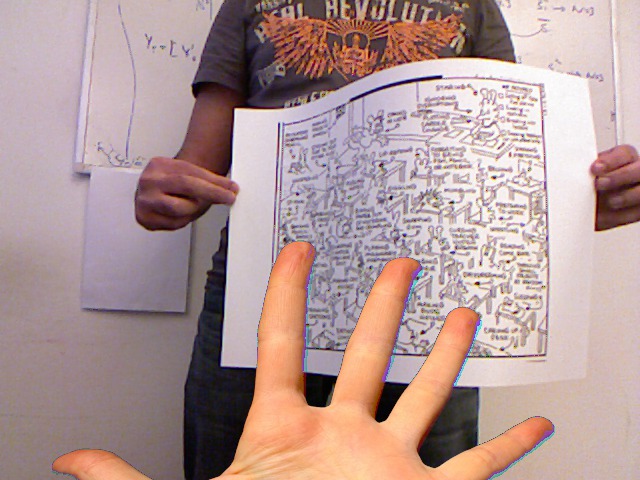}
\includegraphics[width=\figureWidth]{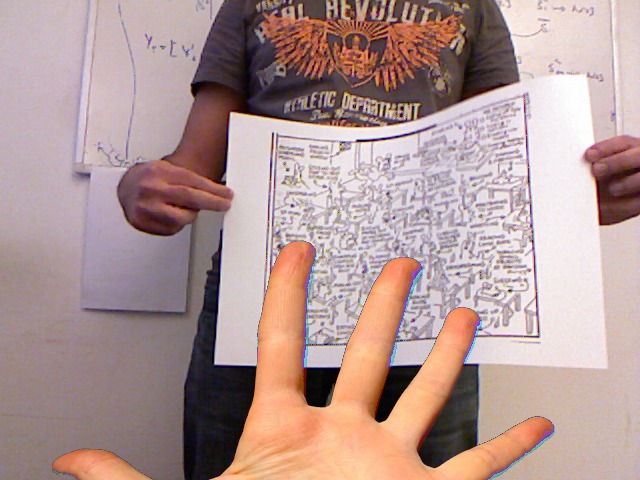}

\includegraphics[width=\figureWidth]{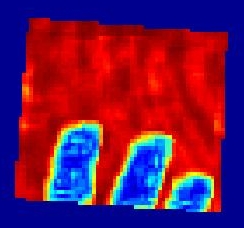}
\includegraphics[width=\figureWidth]{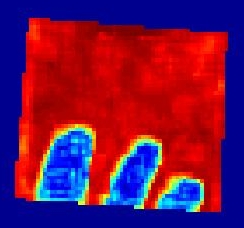}
\includegraphics[width=\figureWidth]{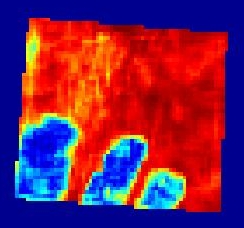}
\includegraphics[width=\figureWidth]{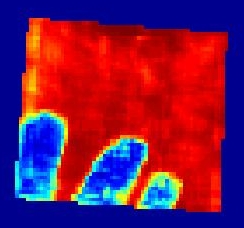}
\includegraphics[width=\figureWidth]{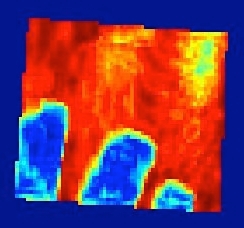}

\vspace{0.1cm}
\caption{Relevancy scores for the well-textured paper dataset.}
\vspace{-0.5cm}
\label{fig:comp_ncc}
\end{figure}

The similarity scores are then normalized to lie in $[0,1]$, and outlier data is also limited at this stage in a process similar to an M-estimator.  The mean $\mu$ and standard deviation $\sigma$ of the NCC scores are found for each frame, and all values further than $3\sigma$ from the mean are clamped to the interval $\mu \pm 3\sigma$.  These values are then linearly rescaled to lie between 0 and 1, and the normalized weights $\hat{\omega}$ are applied to the data in the image energy term of Equation \eqref{eq:totalEnergy}:
\begin{equation}
E_{\text{image}}(\V) = \sum_{\x} \hat{\omega}(\x) d\big(\phi_T(\mathbf{x}),
  \phi_I(\mathbf{W(x;V)}) \big) ,
 \label{eq:energy}
\end{equation}
where the sum here is extended to all the pixels of the template. Relevancy scores for the well-textured paper dataset are shown in Fig.~\ref{fig:comp_ncc}.

\subsection{Handling Sparsely Textured Surfaces}

The relevancy score described in Section~\ref{sec:newMethod} is also able to handle sparsely  textured surfaces.  Image regions containing little or no texture have low relevancy scores, so these pixels will not negatively influence the image alignment.  For example, see Fig.~\ref{fig_relevancy_scores_low_texture}.  Using the proposed relevancy score to weight the utility of the image information coming from each pixel in the image allows the optimization to be driven by the most meaningful available information.  





\section{Experiments and Results}
\label{sec:experiments}
3D surface reconstructions are computed with and without occlusion on both well and poorly textured deforming surfaces. We compare recent methods described in Section \ref{sec_related_work}, which are representative of the current state-of-the-art, against our dense template matching-based reconstruction methods using the various similarity measures and occlusion handling techniques described in Section \ref{sec_framework}.

In particular, we report detailed results of comparisons with the following methods: ``Bartoli12'' \cite{Bartoli12b}, that reconstructs the surface by analytically solving a system of PDEs starting from an estimated 2D parametric warp between images; ``Chhatkuli14'' \cite{Chhatkuli14}, that infers the surface shape exploiting the depth gradient non-holomonic solution of a PDE; ``Brunet14'' \cite{Brunet14}, that reconstructs a smooth surface imposing soft differential constraints of isometric deformation; ``Ostlund12'' \cite{Ostlund12}, that introduces the Laplacian mesh smoothing we employ; and ``Salzmann11'' \cite{Salzmann11a}, that uses pre-learned linear local deformation models.\footnote{Code provided by the authors of these papers was used for all comparisons.}

As for pixel-based template matching techniques, comparing pixel intensity values ``Intensity'' and gradient direction values ``GD'' are done using \ac{SSD}. We also compare standard ``NCC'' and ``MI'' over intensity values.  The ``GBDF'' features are compared using \ac{SSD}, and were seen to be the strongest feature descriptor, so it is these values that we test in the M-estimator framework using the ``Huber'' and ``Tukey'' loss functions.  Our proposed framework from Section \ref{sec:newMethod} is labeled ``GBDF+Oc'' in the figures.  We see that it achieves state-of-the-art performances on a standard, well-textured dataset, and it achieves optimal reconstruction performance in all datasets with occlusions and low texture.

Image sequences were acquired using a Kinect camera, and ground truth surfaces were generated from the depth information. The template is constructed from the first frame, and 3D reconstruction is performed for the rest of the sequence using the image information alone.  The initial mesh coordinates for each frame are set to the locations of the final reconstruction of the previous frame in the sequence.

We consider two different metrics to define the reconstruction accuracy. Many previous methods compare the average distance of the reconstructed 3D mesh vertices to their closest projections onto the depth images. This metric ignores the correspondences between the mesh points and the point cloud.  As a more meaningful metric, we use the Kinect point cloud to build ground truth meshes, and compute the average vertex-to-vertex distance from the reconstructed mesh to the ground truth mesh. This metric is used for the paper itself. Results using the vertex-to-point-cloud distance are provided in the supplementary material.

To ensure a fair comparison, all results are presented using the best parameter values found for each method, tuned separately. To ensure that our results are not overly sensitive to the selection of parameters $\lambda_L$ and $\lambda_S$, we performed the full reconstruction on the well-textured paper dataset over a wide range of values, as presented in Table~\ref{tab:table_diff_lamda}.  It can be observed that increasing or decreasing these parameters by a factor of two around $\lambda_L = 1$ and $\lambda_S = 0.25$ results in very little change in the final reconstruction accuracy, implying that the method is sufficiently insensitive to these parameters as long as they are within a reasonable range.

The surface rest shape is modeled by a $10\times13$ triangle mesh in the well-textured dataset, $14 \times 17$ in the sparsely textured dataset, and  $15\times14$ on the T-shirt dataset.  The $\sigma$s used in the hierarchical procedures were $\{15,7,3\}$ and $\{5,3,2\}$.

\begin{table}
\vspace{-0.2cm}
\caption{Reconstruction errors over a range of weighting coefficient values using the well-textured paper dataset.}
\hspace{0.2cm}
\setlength{\tabcolsep}{4pt}
\centering
\scriptsize
\begin{tabular}{|c|c|c|c|c|c|c|c|c|c|}
       \hline
\multicolumn{2}{|c|}{\multirow{2}{*}{ error (mm) } }
      &  \multicolumn{8}{c|}{$\lambda_L$} \\
       \cline{3-10}
\multicolumn{2}{|c|}{}
     & 10 & 2 & 1 & 0.5 & 0.25 & 0.1 & 0.05 & 0.01 \\
      \hline
     \multirow{8}{*}{ \begin{sideways} $\lambda_S$\end{sideways} } 
      & 10    & 7.46 & 6.46 & 5.73 & 5.59 & 5.31 & 18.45 & 93.07 & N.A \\ \cline{2-10}
      & 2     & 4.60 & 1.58 & 2.41 & 3.68 & 4.49 & 5.17 & 17.23 & 319.09 \\ \cline{2-10}
      & 1     & 1.93 & 1.39 & 1.20 & 1.97 & 3.44 & 4.61 & 5.94 & 147.96 \\ \cline{2-10}
      & 0.5   & 2.02 & 1.73 & 1.43 & 1.08 & 1.80 & 3.76 & 4.77 & 61.87 \\ \cline{2-10}
      & 0.25  & 2.01 & 1.86 & {\bf 1.67} & 1.45 & 1.10 & 2.17 & 3.76 & 26.18 \\ 
      \cline{2-10}
      & 0.1   & 2.05 & 1.91 & 1.75 & 1.62 & 1.57 & 1.20 & 1.68 & 240.66 \\ \cline{2-10}
      & 0.05  & 2.07 & 2.03 & 1.96 & 1.86 & 1.82 & 5.62 & 14.86 & 185.47 \\ \cline{2-10}
      & 0.01  & 18.23 & 15.44 & 6.25 & 6.45 & 6.31 & 10.44 & 14.11 & 342.12\\ 
        \hline
\end{tabular}
\label{tab:table_diff_lamda}
\vspace{-0.5cm}
\end{table}

Our  approach  relies on  frame-to-frame  tracking  and thus  requires  a
  sufficiently good initialization. However, because the method has a wide basin
  of convergence, a rough initialization suffices.  Our method can fail when the
  initialization is too far from  the solution, when frame-to-frame deformations
  are so  large that  the relevancy  scores stop being  reliable, or  when large
  changes in  surface appearance  and severe occlusions  cause the  image energy
  term  to  become  uninformative.   If  the   tracking  is  lost,  it  must  be
  reinitialized, for  example by  using a feature  point based  method. However,
  this did not prove to be necessary to obtain any of the results shown below.

\subsection{Basin of Convergence}  
\label{basin_of_convergence}


\begin{figure*}
\vspace{-3mm}
\centering
\newcommand{\figureWidth}{2.1cm}
\newcommand{\figureHeight}{1.5cm}
{\subfigure[] { \includegraphics[width=2.1cm,height=\figureHeight]{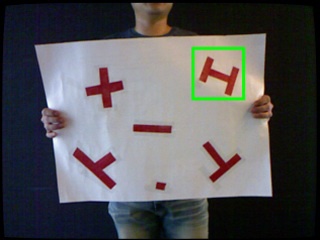} }} \hspace{2mm}
{\subfigure[] { \includegraphics[width=3cm,height=\figureHeight]{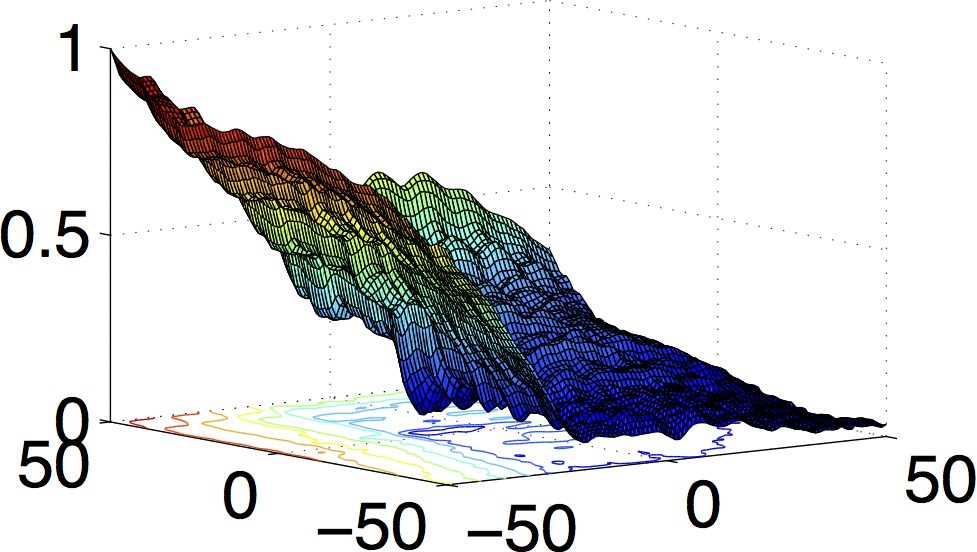} }} \hspace{2mm}
{\subfigure[] { \includegraphics[width=3cm,height=\figureHeight]{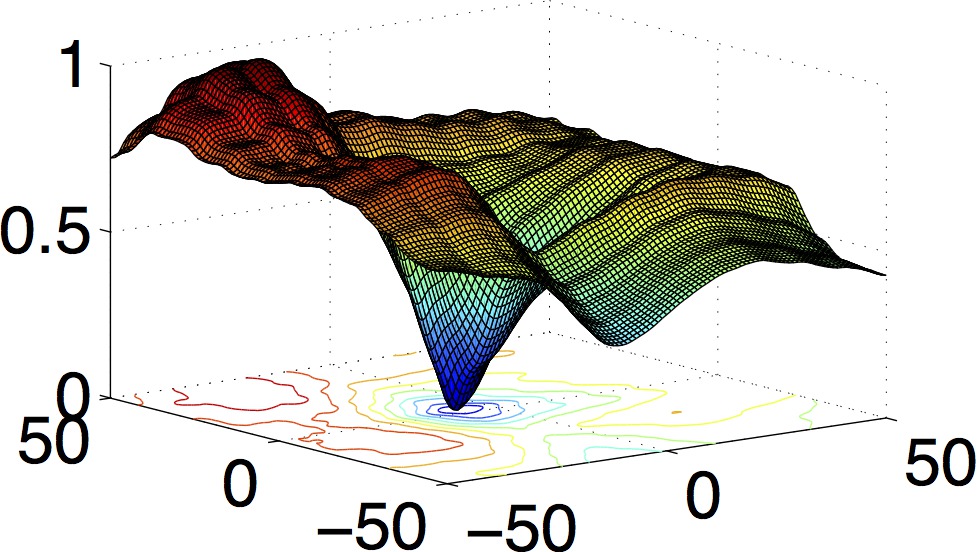} }} \hspace{2mm}
{\subfigure[] { \includegraphics[width=3cm,height=\figureHeight]{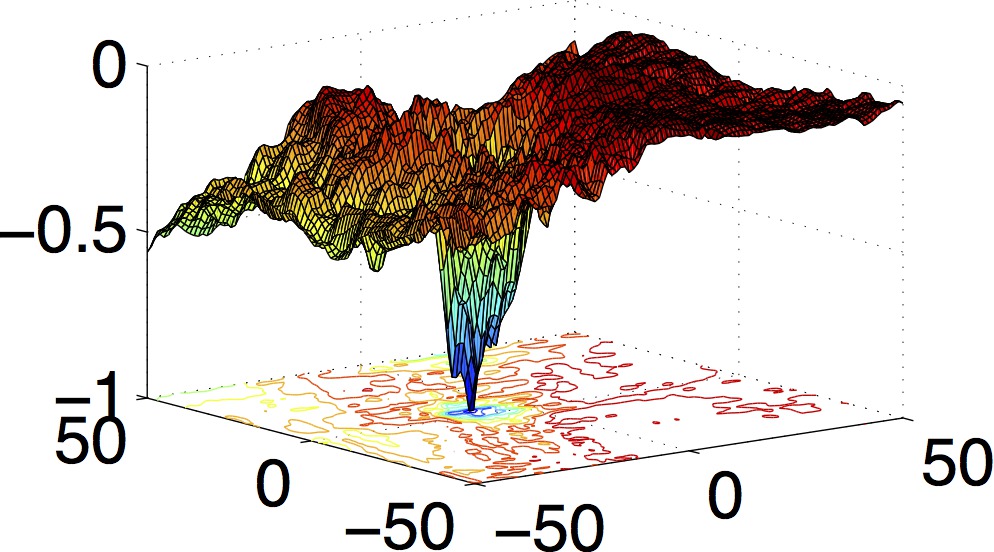} }} \hspace{2mm}
{\subfigure[] { \includegraphics[width=3cm,height=\figureHeight]{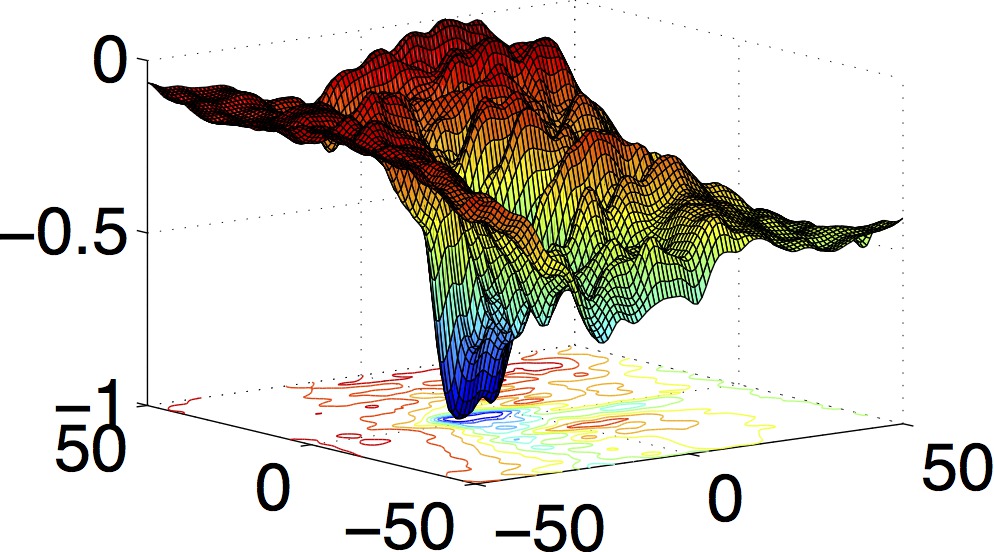} }} \\ \vspace{-4mm}
{\subfigure[] { \includegraphics[width=2.1cm,height=\figureHeight]{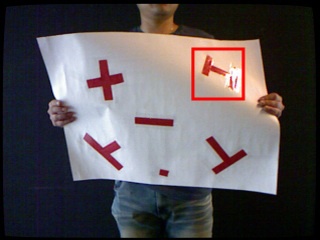} }} \hspace{2mm}
{\subfigure[] { \includegraphics[width=3cm,height=\figureHeight]{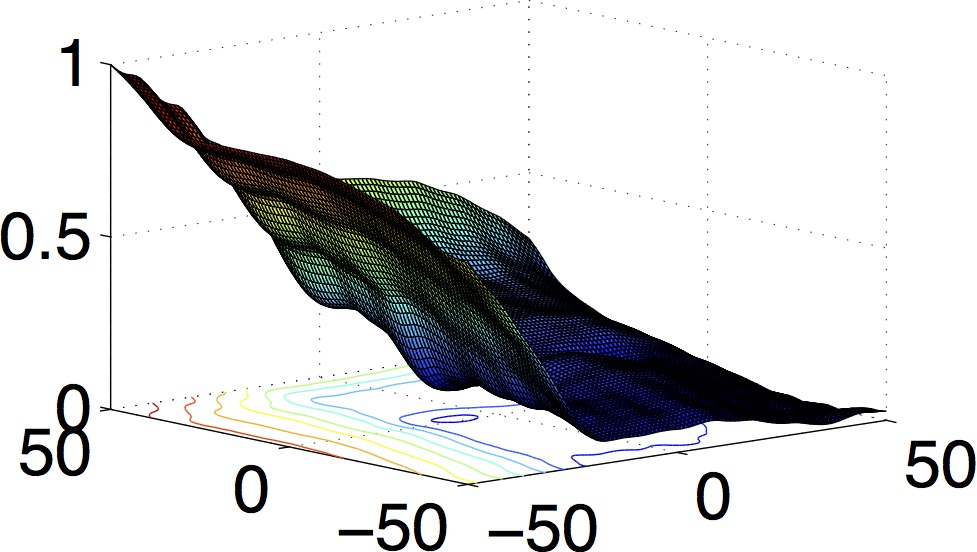} }} \hspace{2mm}
{\subfigure[] { \includegraphics[width=3cm,height=\figureHeight]{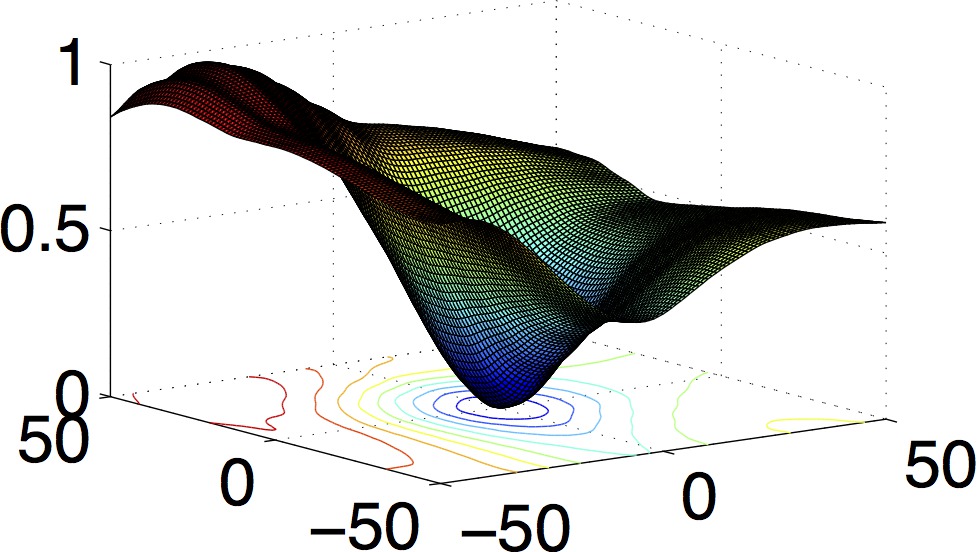} }} \hspace{2mm}
{\subfigure[] { \includegraphics[width=3cm,height=\figureHeight]{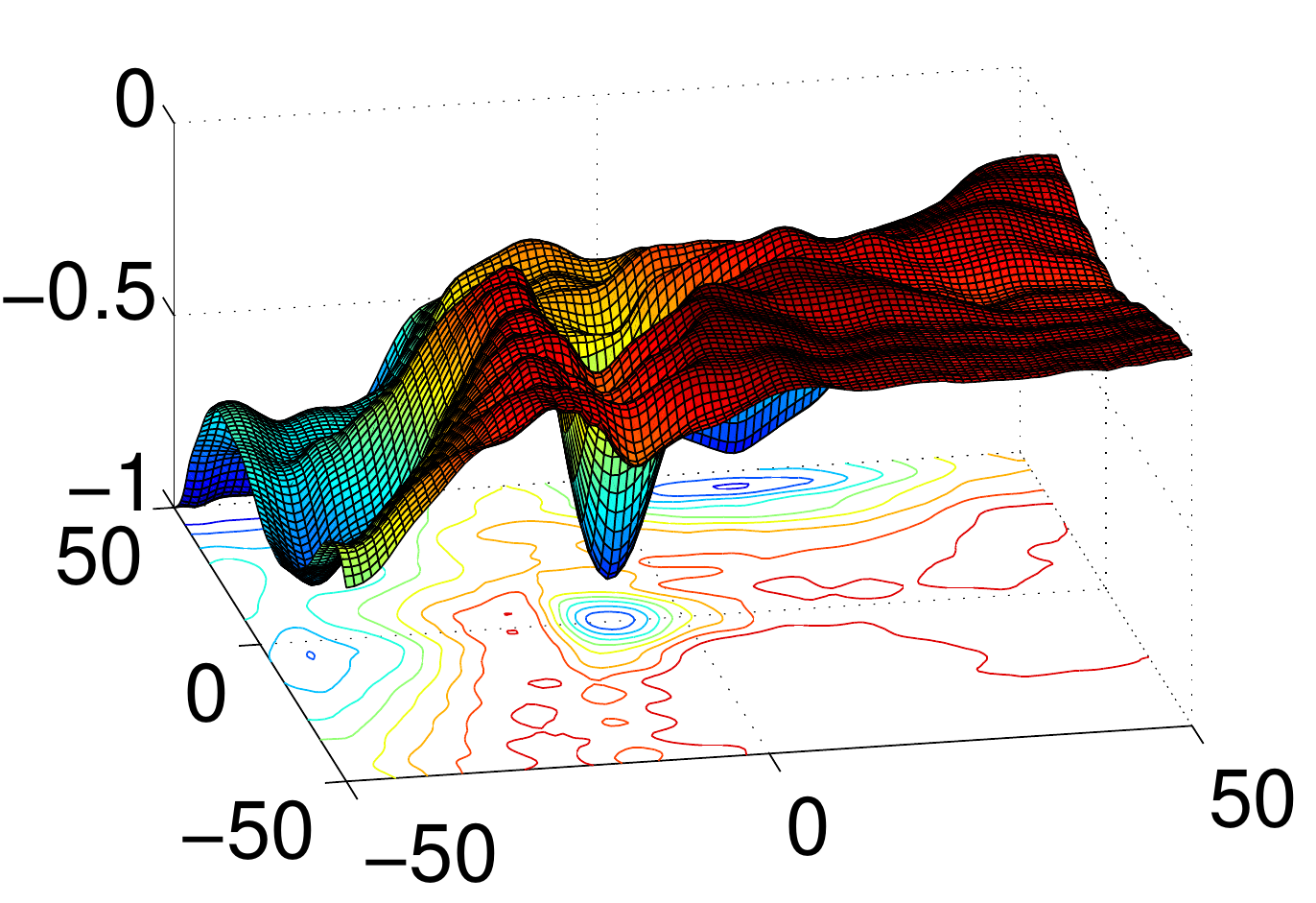} }} \hspace{2mm}
{\subfigure[] { \includegraphics[width=3cm,height=\figureHeight]{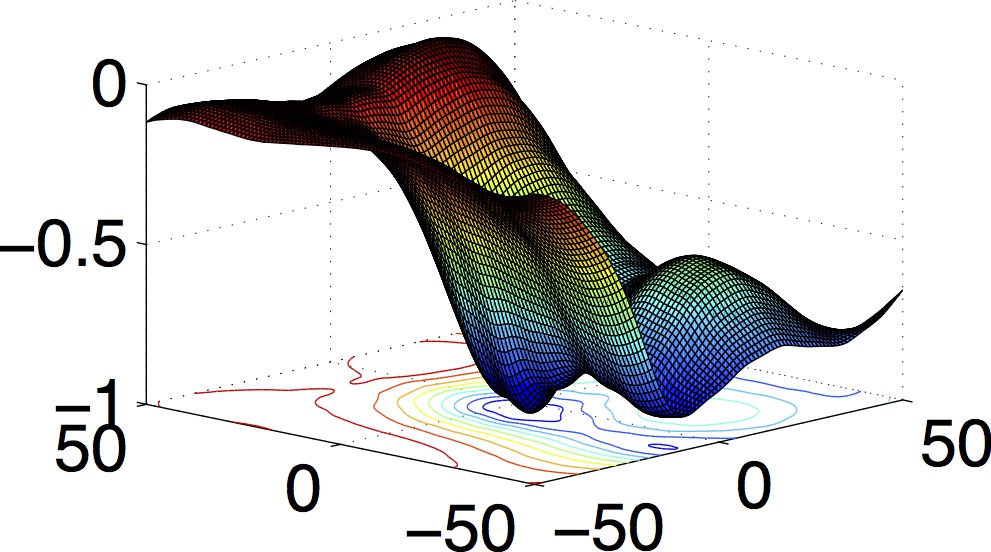} }}
\vspace{-1mm}

\caption{Robustness of alignment functions {\it w.r.t.} translations between (a) template, and (f) input image, showing the basin of convergence of the alignment costs around the correct position using {\bf Top row:} weak Gaussian smoothing, {\bf Bottom row:} strong Gaussian smoothing, over (b)(g) Intensity, (c)(h) GBDF, (d)(i) MI, (e)(j) NCC.}
\label{fig:exp_align}
\vspace{-1mm}
\end{figure*}

To understand the limitations of the various cost functions, we conducted a simple alignment experiment to test their respective sensitivities to initial position and image distractors; results are presented in Fig.~\ref{fig:exp_align}.  The red image window in the input image is translated in x and y about the known best alignment to the green template window, and the cost to compare each window pair is plotted, to allow the basins of convergence to be inspected visually. \ac{MI} and \ac{NCC} both reach a maximum value close to 1 at the point of best alignment, but we invert these functions so that a minimum cost of all functions is expected at the point of best alignment.

The GBDF descriptors from Eq.~\ref{MC_eqn} are seen to have a strong minimum at the point of best alignment, with a reasonably wide, smooth basin of convergence, the desired property of a good cost function.  However, Intensity, MI, and NCC all have several nearby local minima.  Mutual Information is further seen to have a very narrow basin of convergence around the correct point of best alignment, meaning that it is very likely to converge to an incorrect alignment given an imperfect initial position.

This experiment only tests translation sensitivity because this is the variation best understood visually, but the similar results are likely from other types of misalignment.

\subsection{Well-Textured Surfaces}

\begin{figure}
  \centering
  \includegraphics[width=0.48\textwidth,height=3cm]{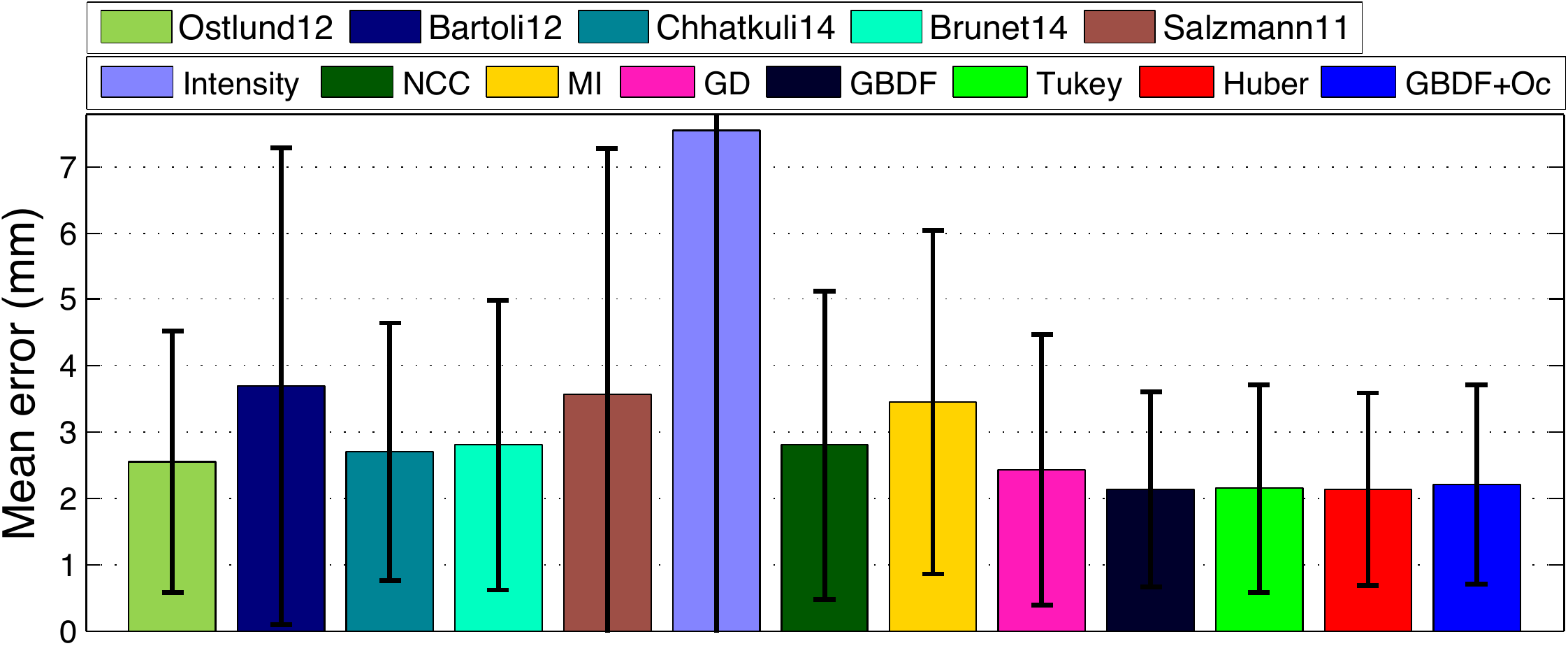}  
  \caption{Reconstruction results on the well-textured paper dataset, no occlusions. All feature-points based methods work reasonably well.}
   \label{fig:capture_paper_errors}
   \vspace{-0.3cm}
\end{figure}

\begin{figure}
\centering
\newcommand{\figureWidth}{2.4cm}
{\subfigure[Ostlund12]
{\includegraphics[width=\figureWidth]{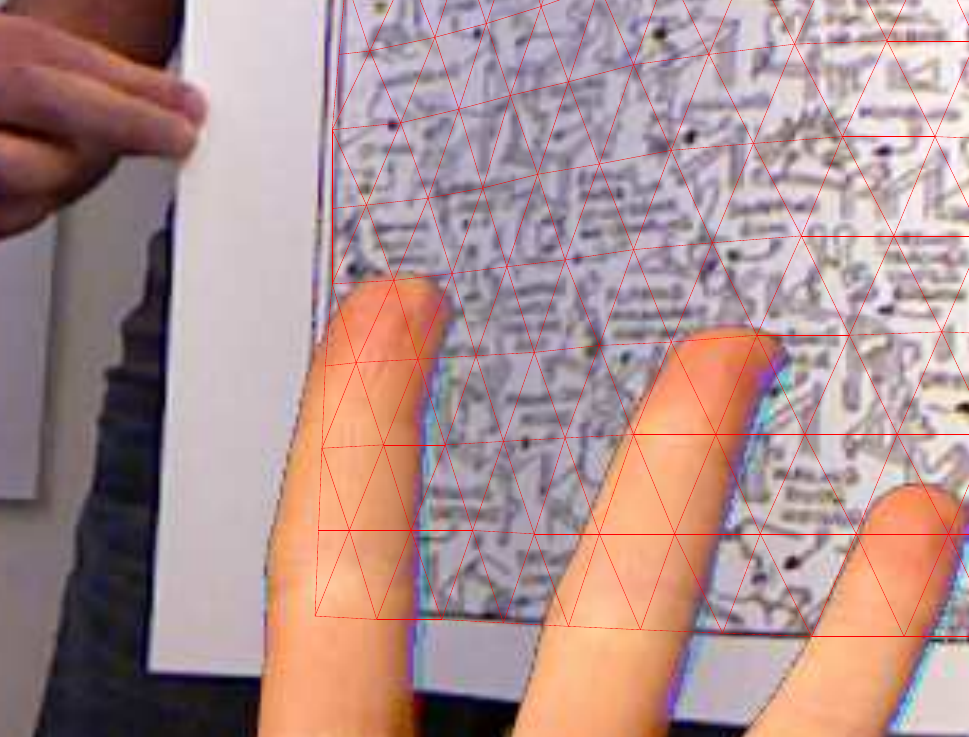}
\hspace{.001cm}}}
{\subfigure[GBDF alone]
{\includegraphics[width=\figureWidth]{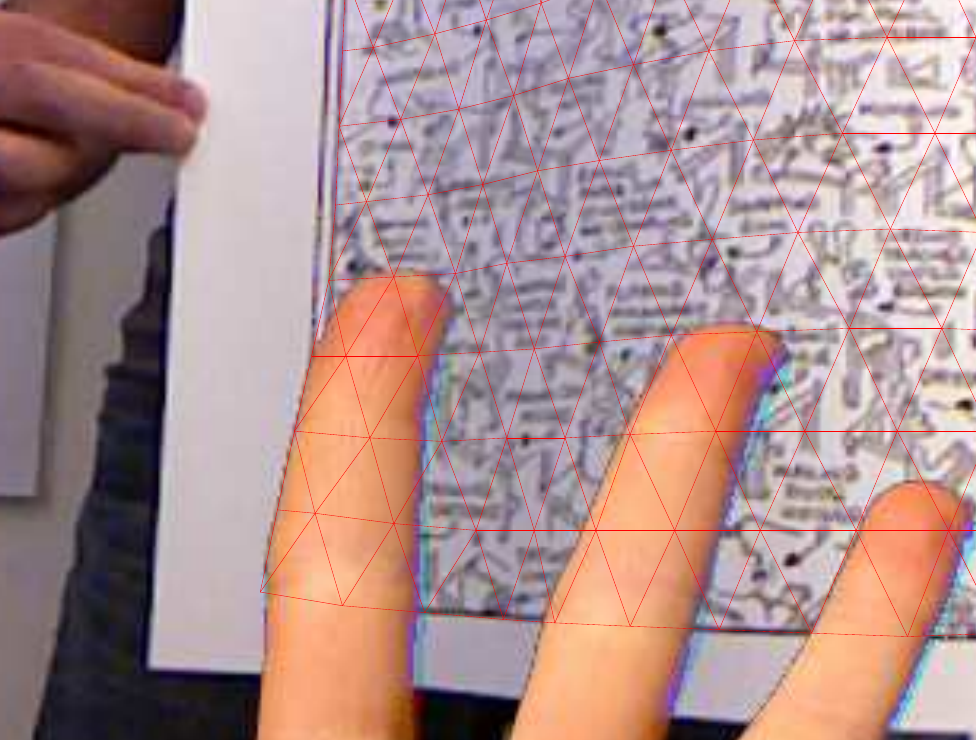}
\hspace{.001cm}}}
{\subfigure[MI]
{\includegraphics[width=\figureWidth]{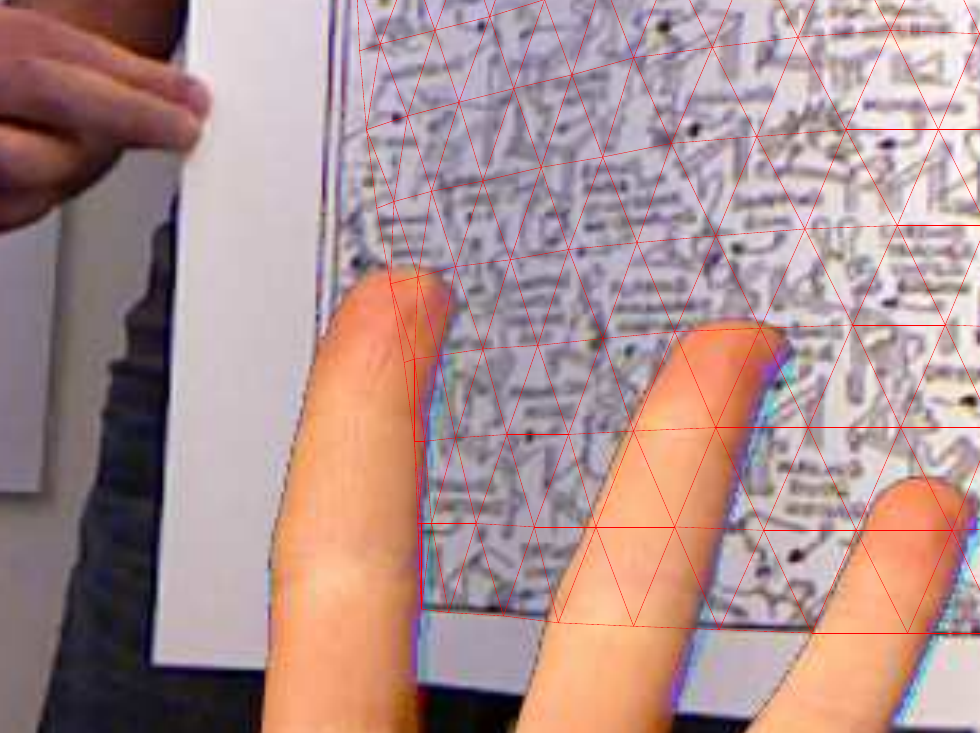}
\hspace{.001cm}}} \\ \vspace{-2mm}
{\subfigure[Tukey] 
{\includegraphics[width=\figureWidth]{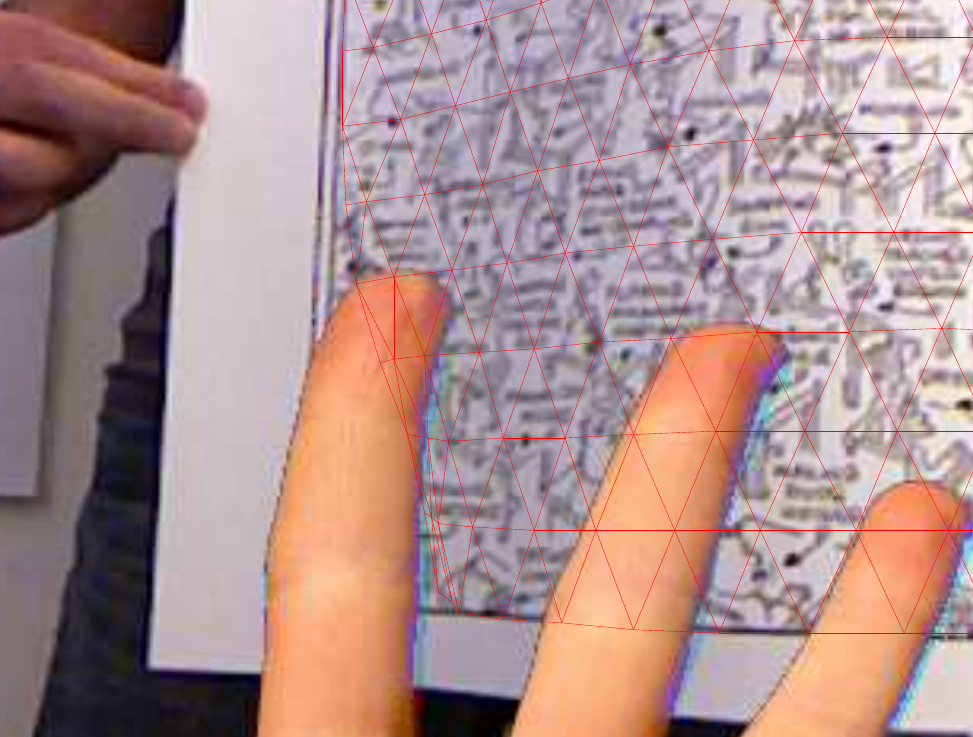}
\hspace{.001cm}}}
{\subfigure[Huber] 
{\includegraphics[width=\figureWidth]{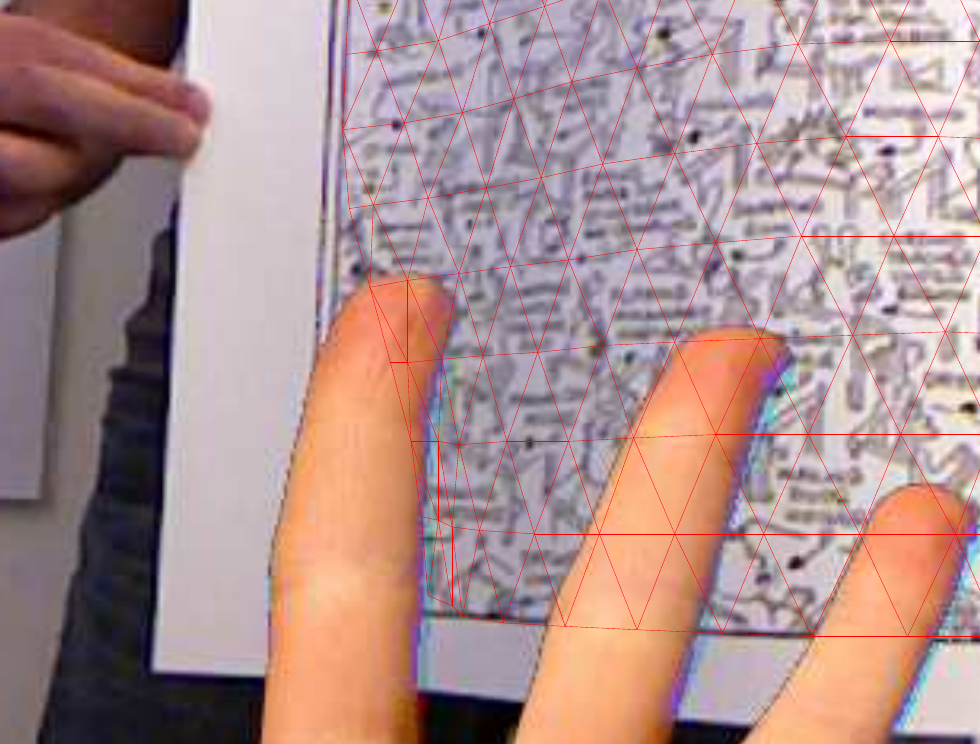}
\hspace{.001cm}}}
{\subfigure[GBDF+Oc]
{\includegraphics[width=\figureWidth]{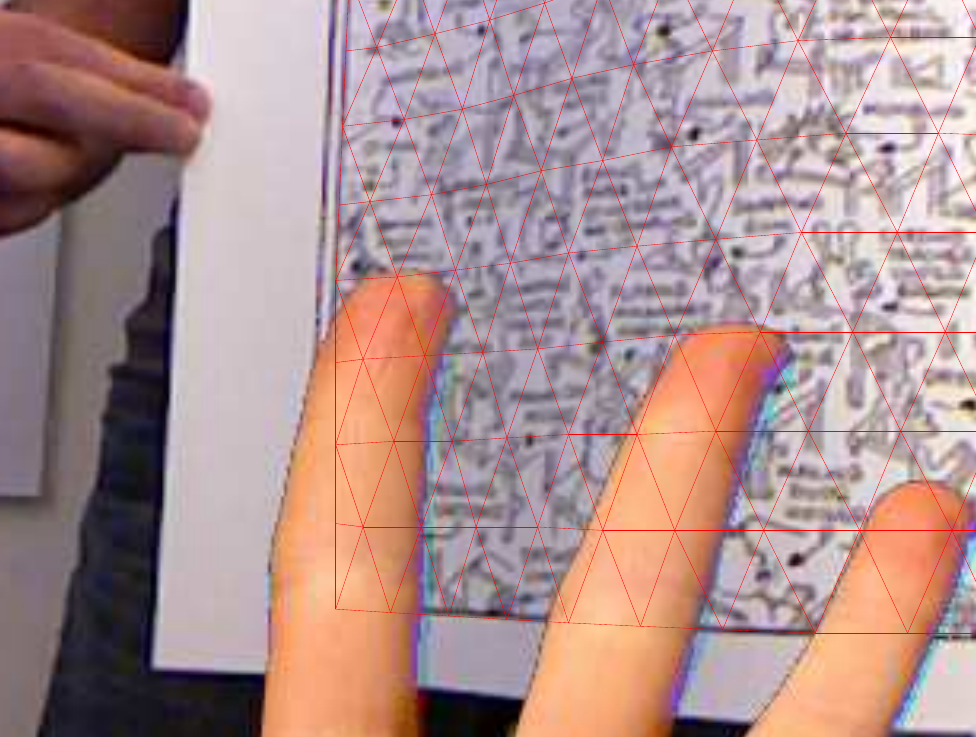} }} \\
\vspace{-1mm}
\caption{Output for a single frame showing relative reconstruction accuracies.  Mutual Information and M-estimators fail to correctly handle the occlusion, while the proposed framework is successful.}
\label{fig:capture_paper_failure}
\vspace{-0.4cm}
\end{figure}

We performed experiments using the well-textured paper dataset presented in \cite{Varol12a} consisting of $193$ consecutive images, for example see Fig.~\ref{fig:capture_paper_failure}.  Quantitative results are presented in Fig.~\ref{fig:capture_paper_errors}.  For this well-textured dataset, all the feature point-based methods work well and dense matching methods are only slightly better.  The biggest errors are due to lighting changes, where intensity features using SSD occasionally fail to track part of the surface, and hence have a higher error.

\begin{figure} 
  \centering
  \includegraphics[width=0.48\textwidth,height=3cm]{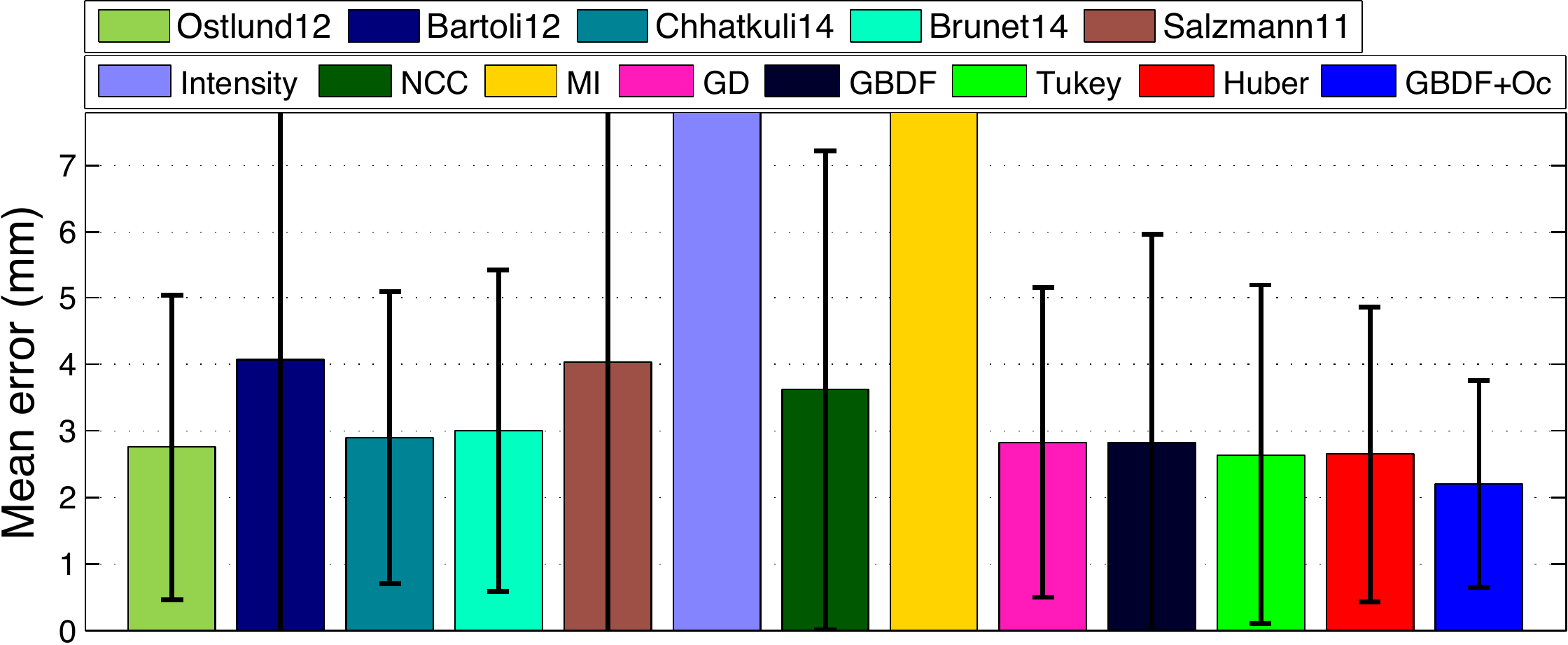}
  \caption{Reconstruction results on the well-textured paper dataset, with occlusions. Feature-based methods are largely robust to occlusion, however the overall depths recovered are not as accurate as the proposed framework that includes occlusion handling.}
  \vspace{-0.4cm}
  \label{fig:capture_paper_oc_errors}
\end{figure}

To evaluate the robustness of each method to occlusion, we add artificial hand image occlusions to the image sequence. The reconstruction results are presented in Fig.~\ref{fig:capture_paper_oc_errors}.
Feature-based methods still produce reasonably good 2D reprojection results in this dataset, but the recovered depths under the occlusion are not very accurate.
Fig.~\ref{fig:capture_paper_failure} provides the output for a single frame where it can be seen that the reconstruction fails when using the strong GBDF without occlusion handling and also when using M-estimators to attempt to handle occlusion, while the proposed framework is still able to able track the surface accurately. In this situation, Mutual Information and both the Tukey and Huber M-estimators are confused by the edges created by the finger and converge to incorrect locations.

\begin{figure*}
\vspace{-0.1cm}
\centering
\newcommand{\figureWidth}{2.0cm}
\includegraphics[width=\figureWidth]{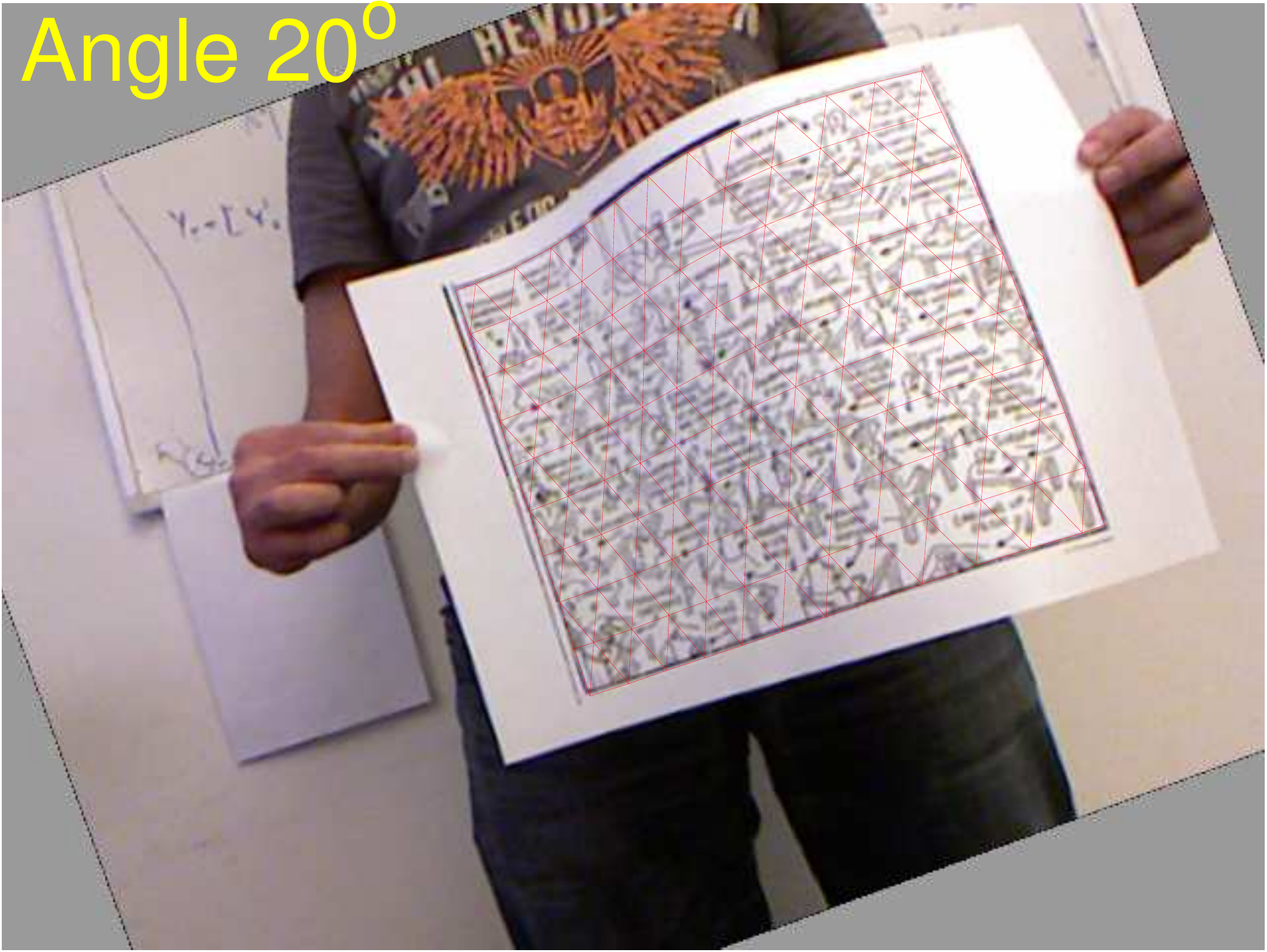} \hspace{2mm}
\includegraphics[width=\figureWidth]{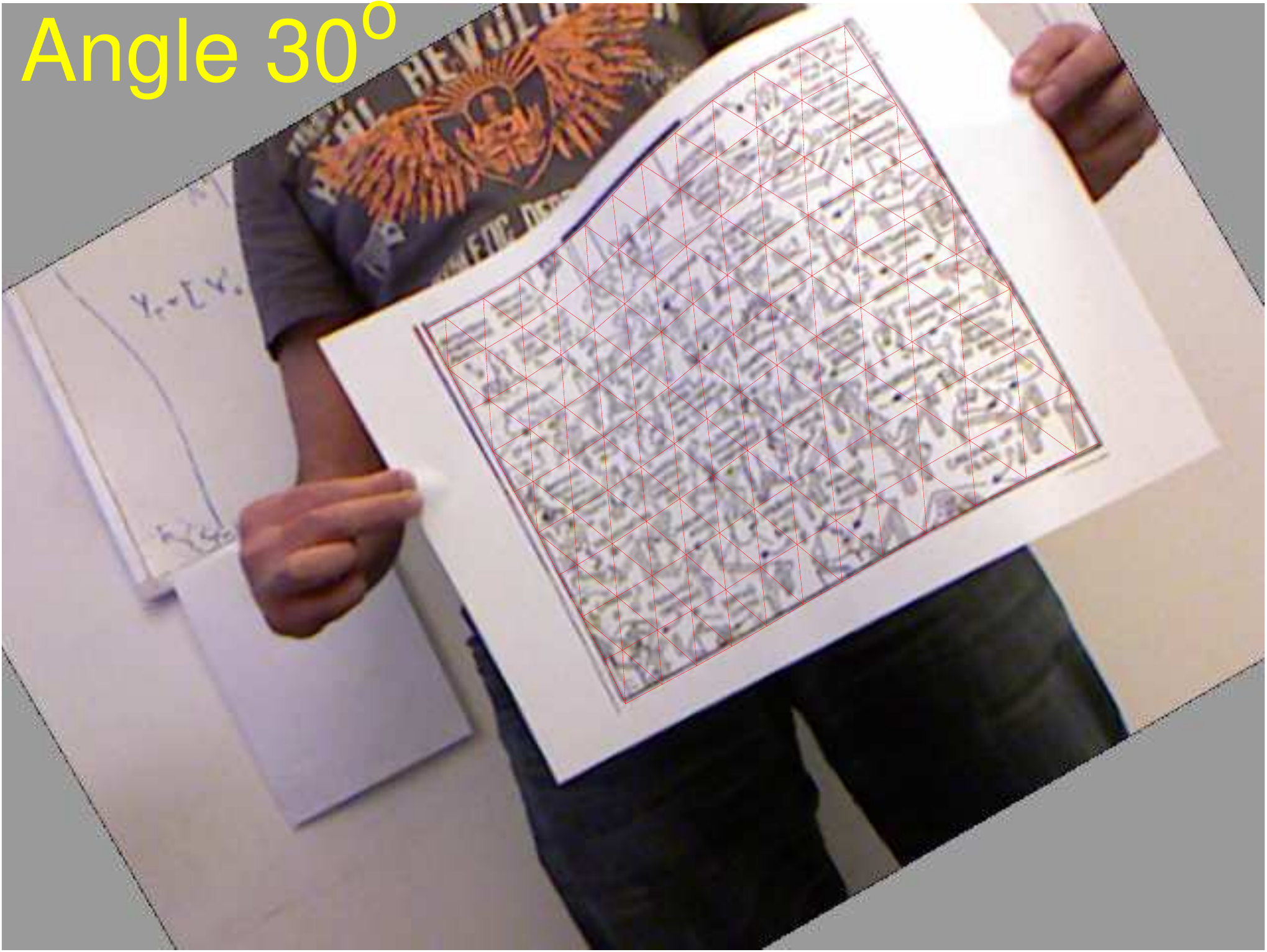} \hspace{2mm}
\includegraphics[width=\figureWidth]{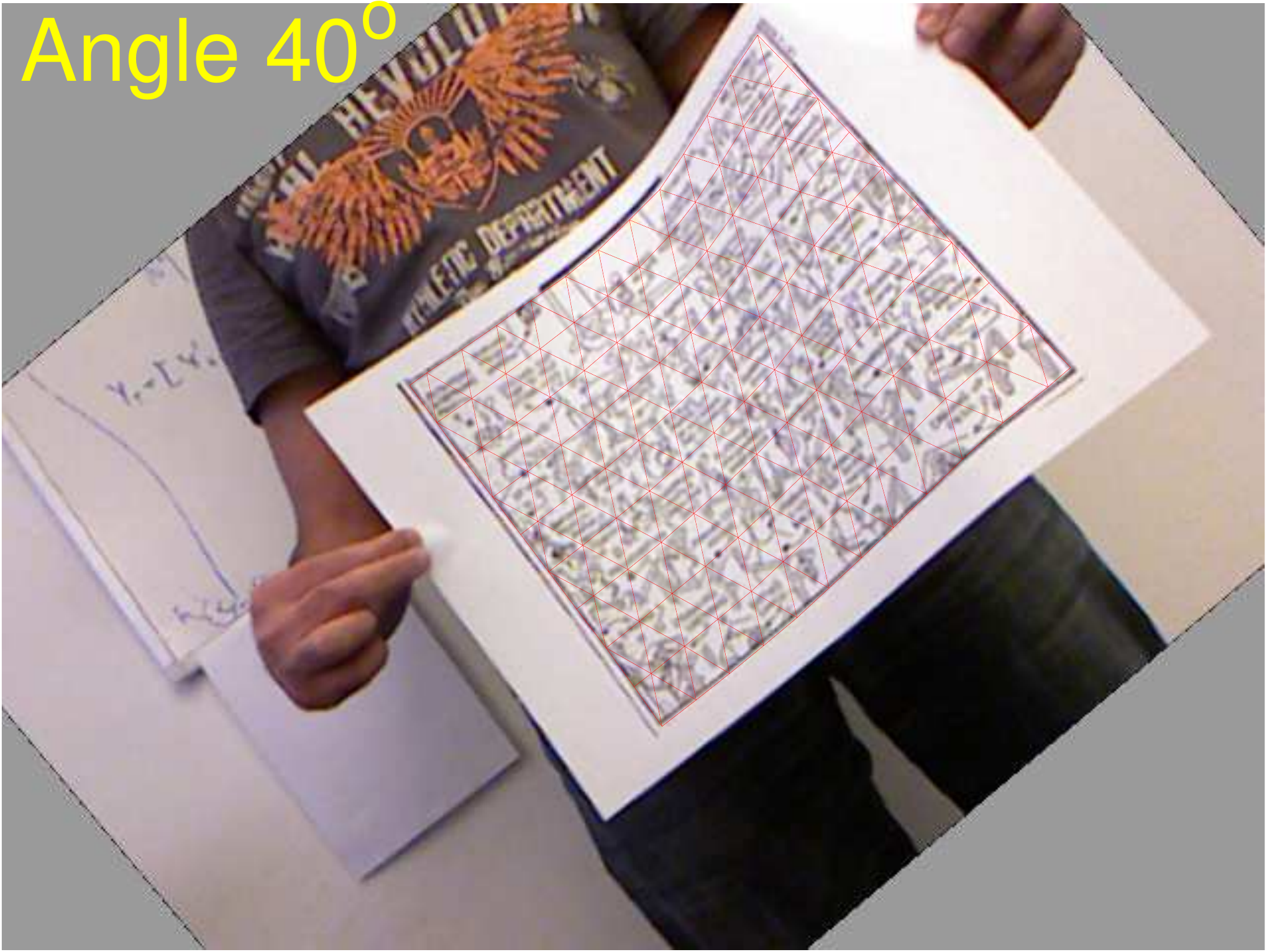} \hspace{2mm}
\includegraphics[width=\figureWidth]{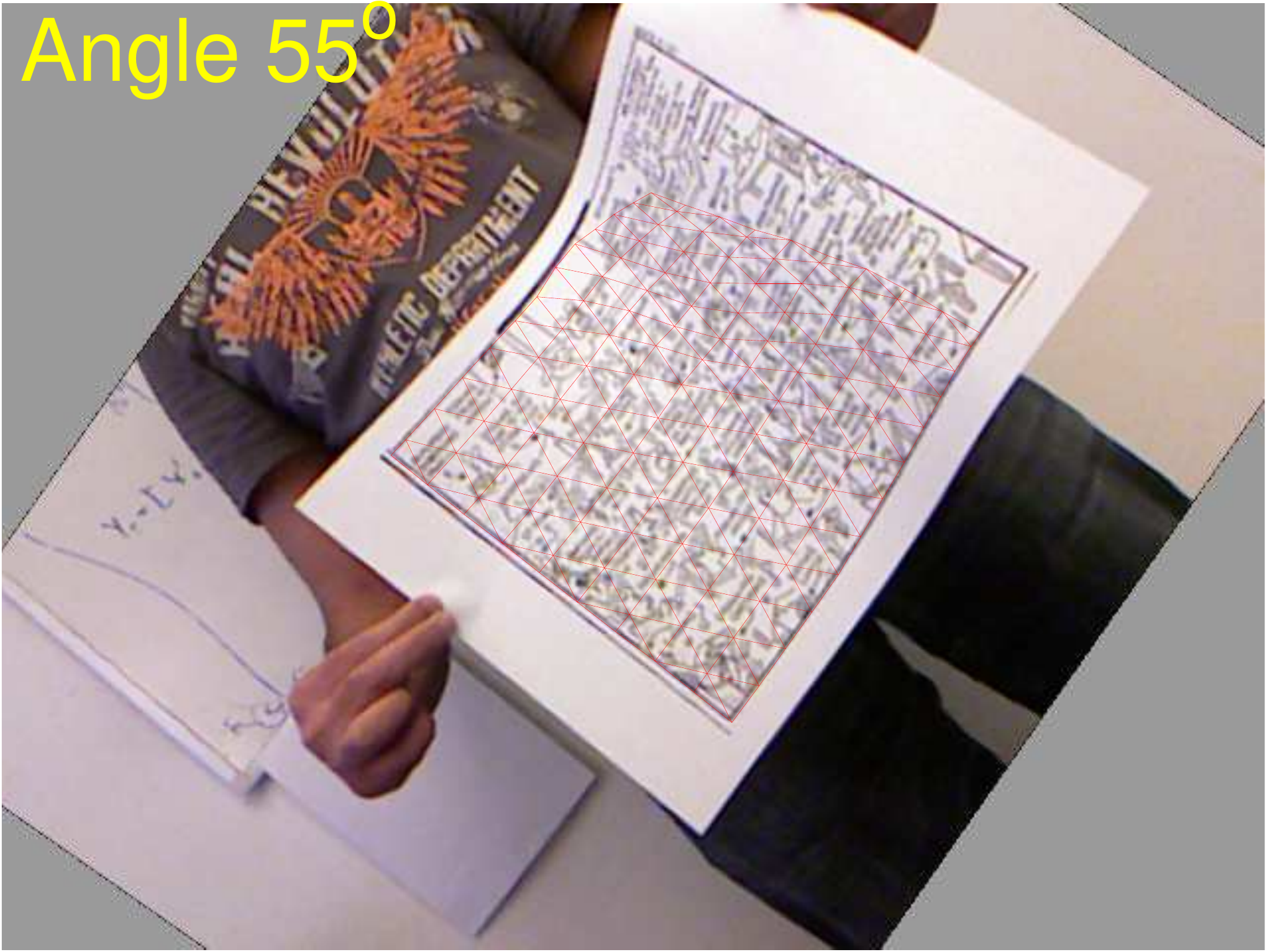} \hspace{2mm}
\includegraphics[width=\figureWidth]{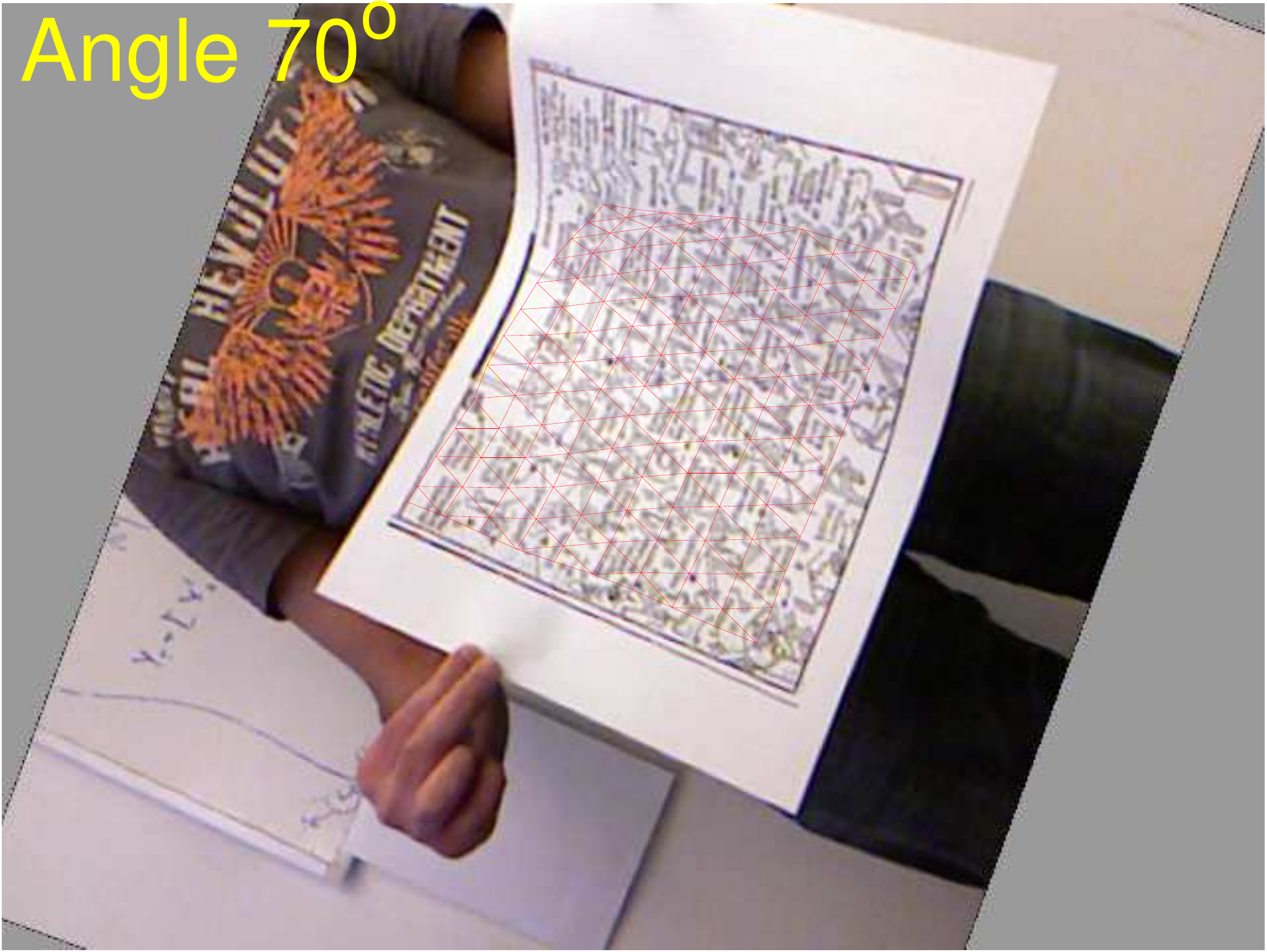} \hspace{2mm}
\includegraphics[width=\figureWidth]{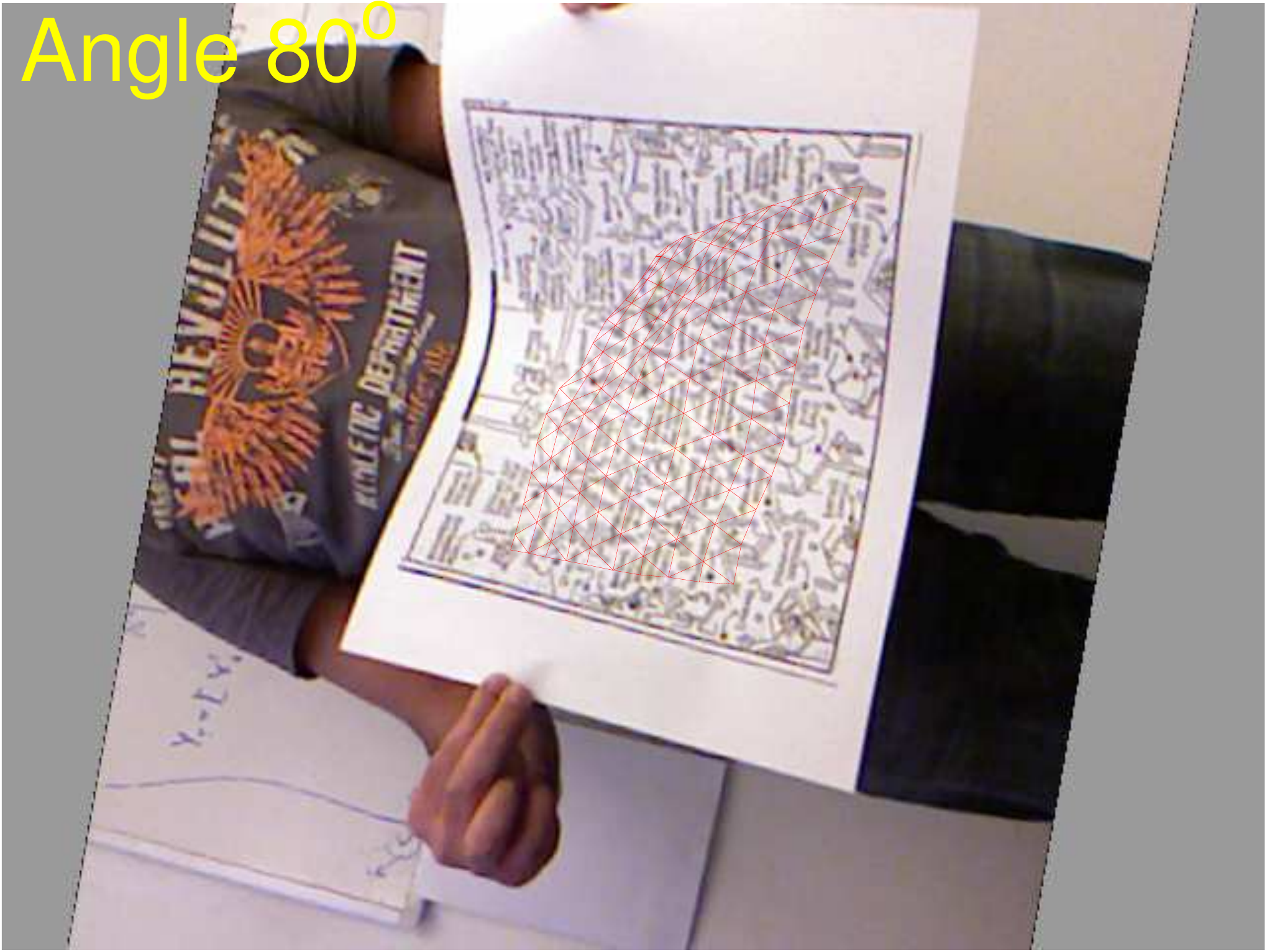} \hspace{2mm}
\includegraphics[width=\figureWidth]{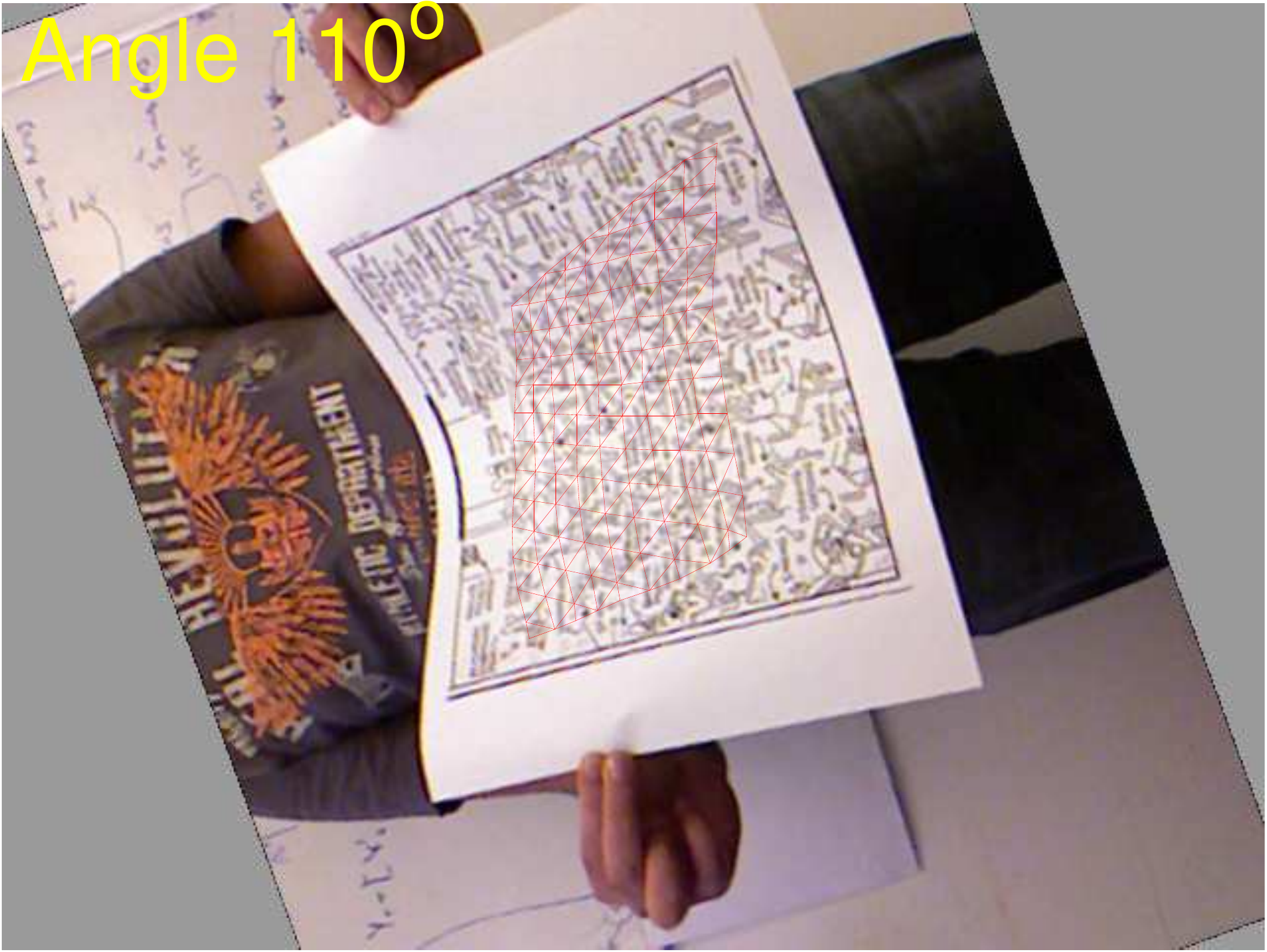}
\\
\hspace{5mm}
\includegraphics[width=\figureWidth]{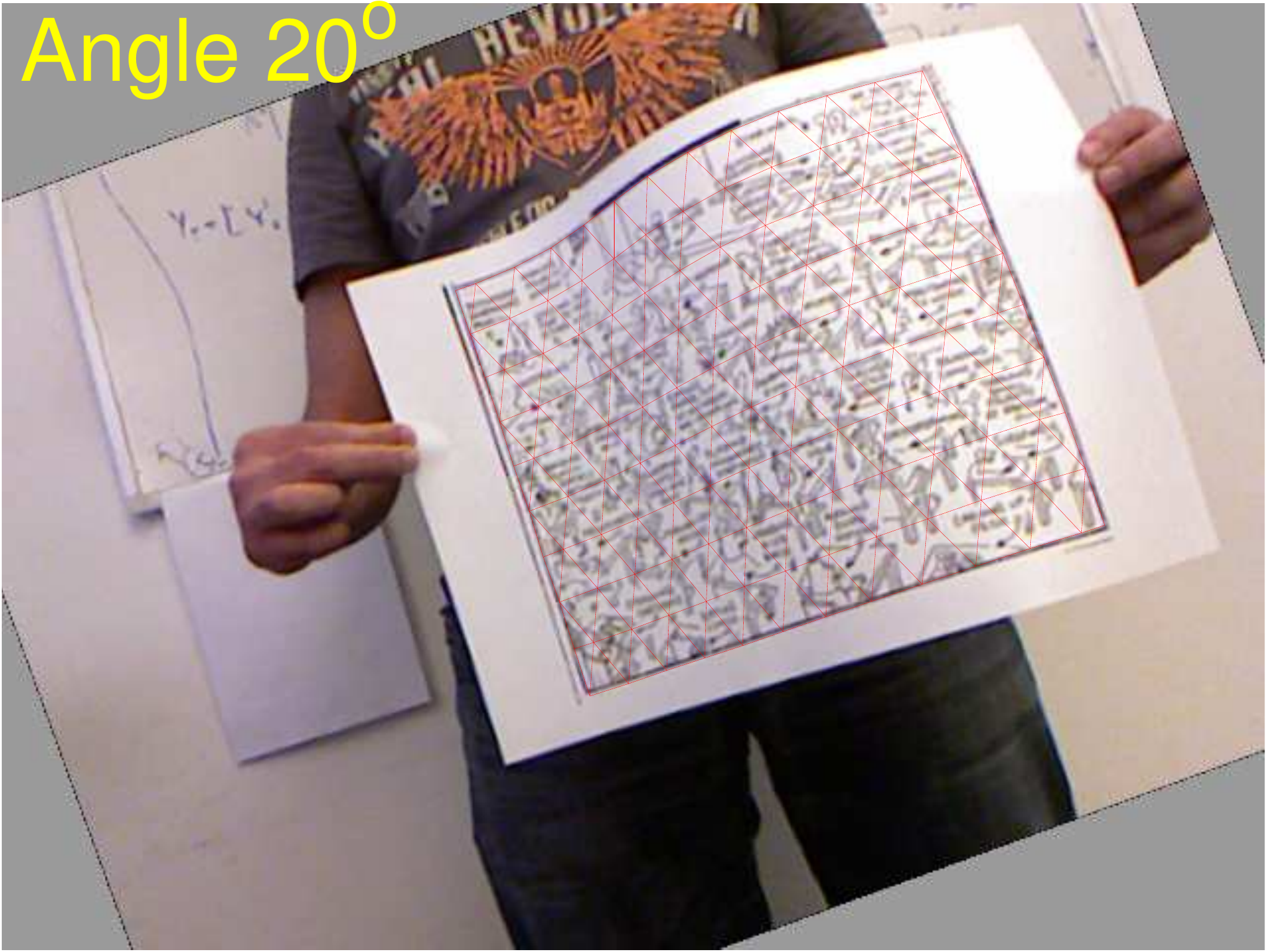} \hfill
\includegraphics[width=\figureWidth]{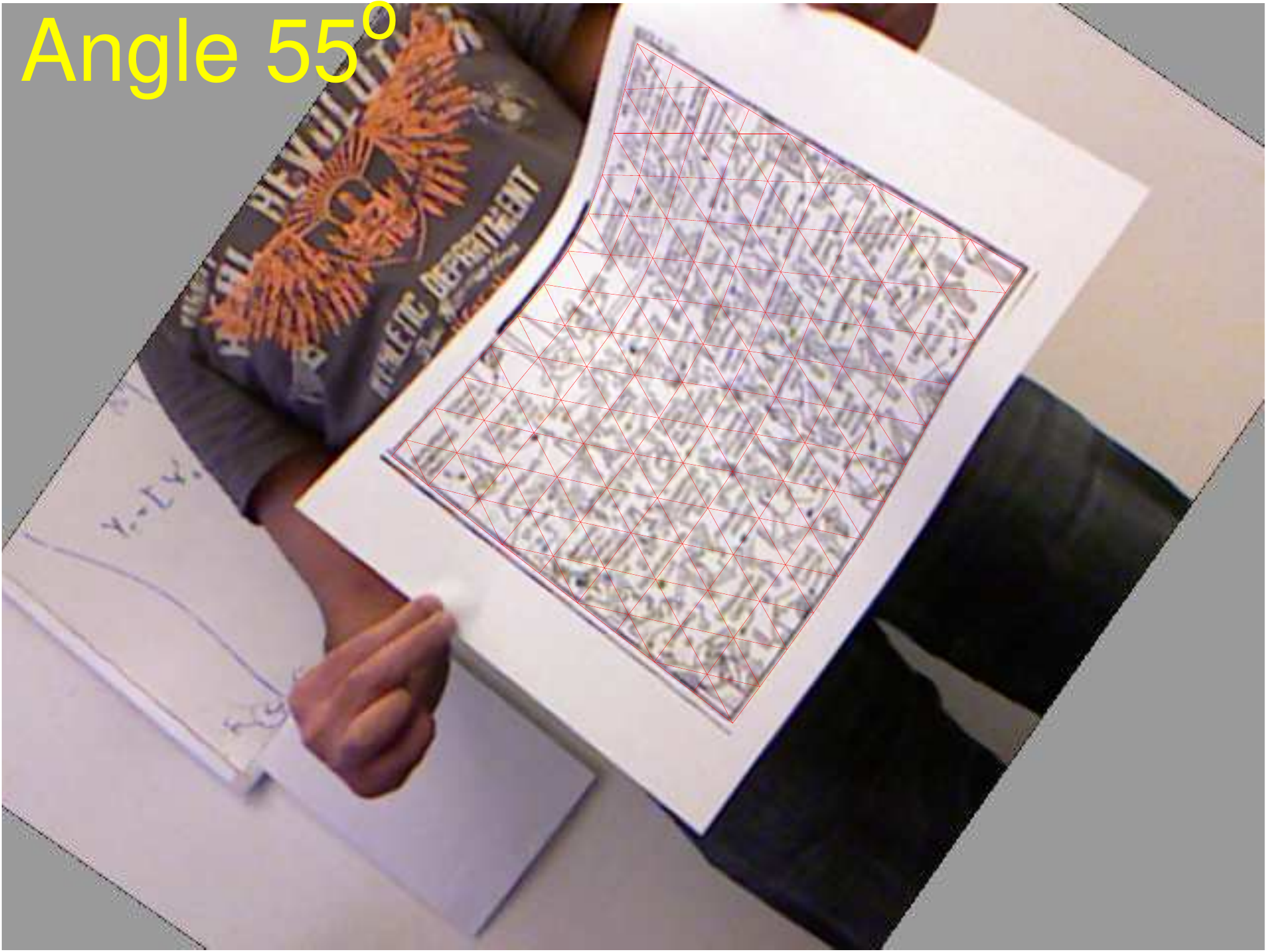} \hfill
\includegraphics[width=\figureWidth]{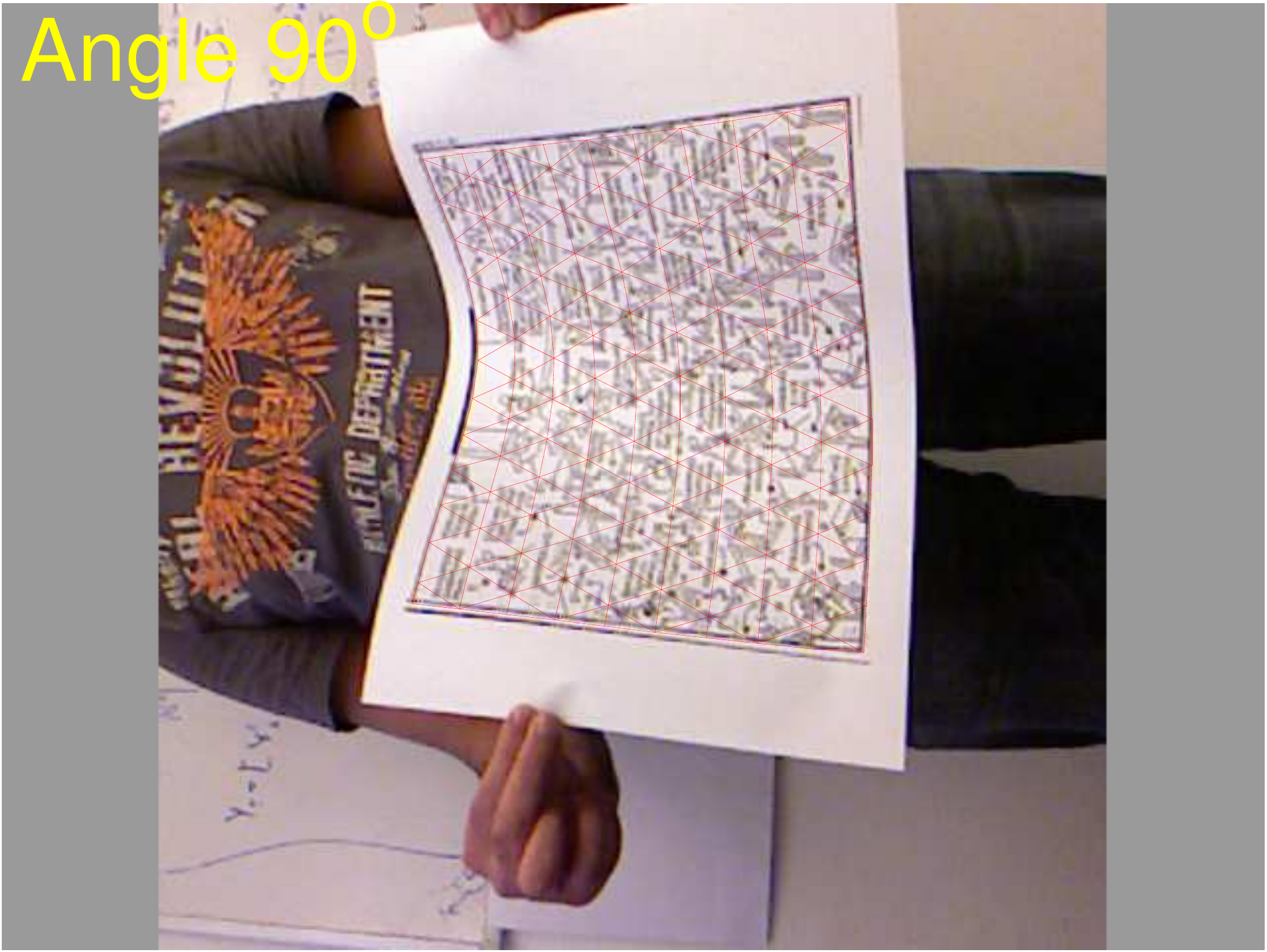} \hfill
\includegraphics[width=\figureWidth]{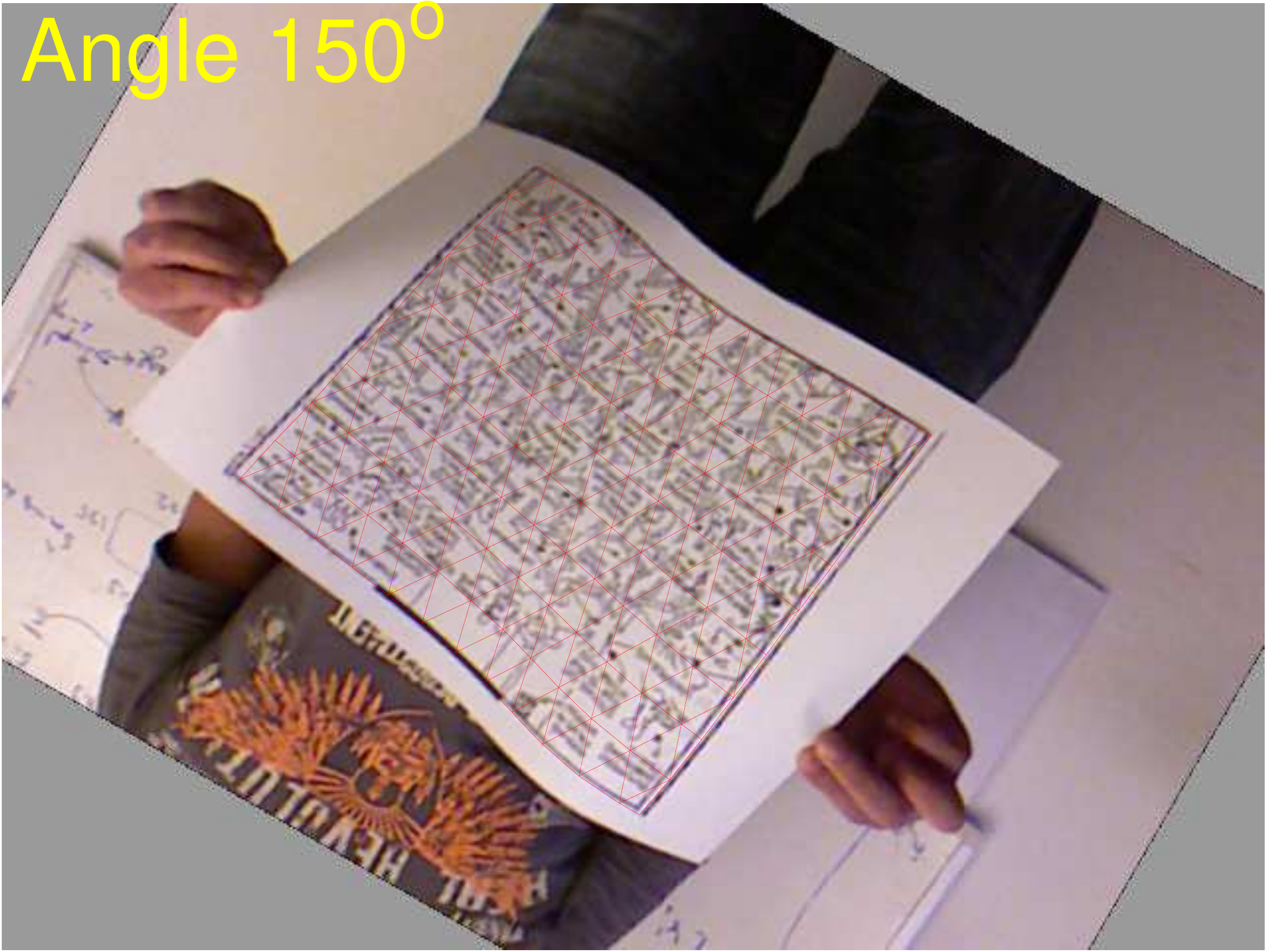} \hfill
\includegraphics[width=\figureWidth]{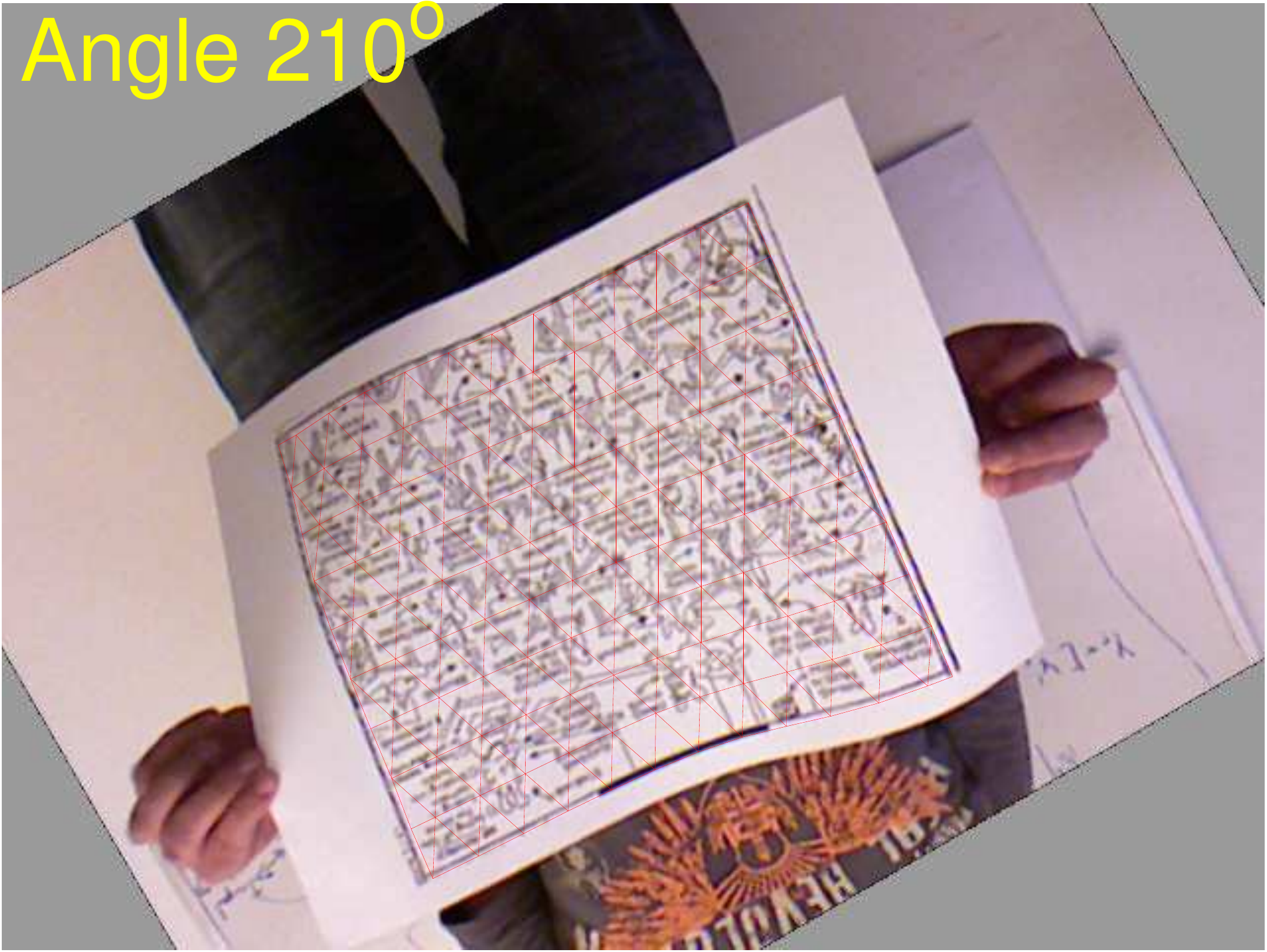} \hfill
\includegraphics[width=\figureWidth]{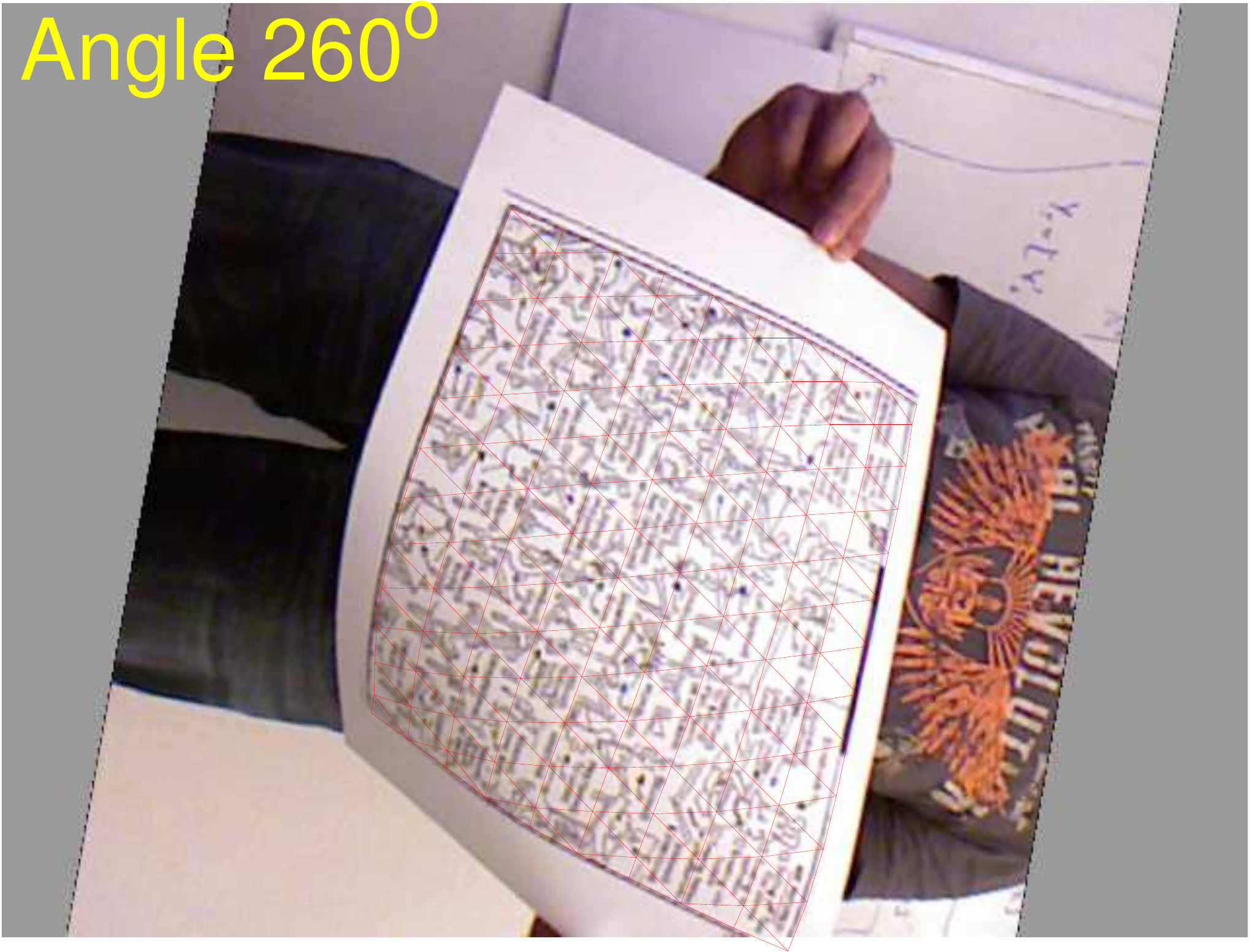} \hfill
\includegraphics[width=\figureWidth]{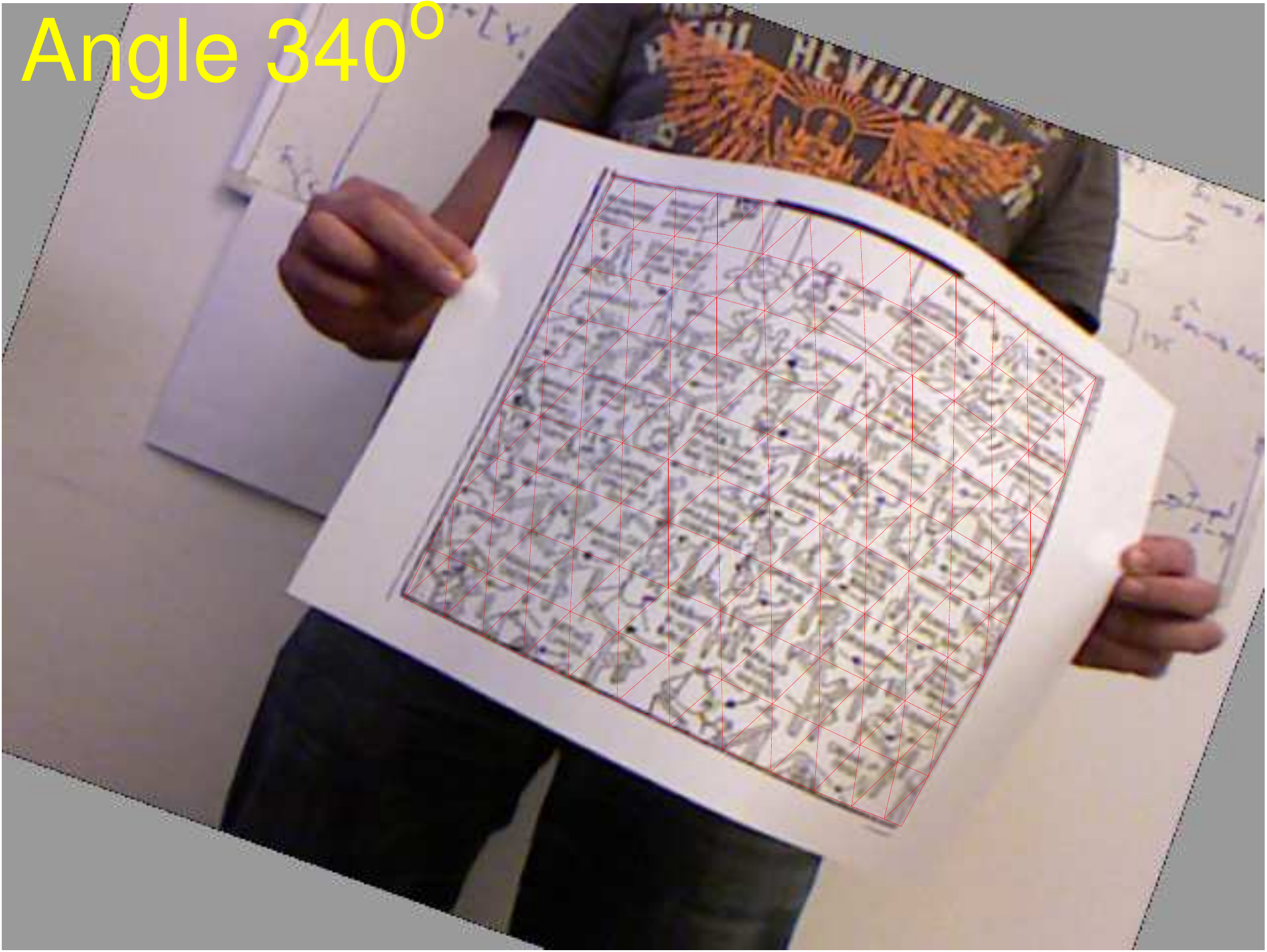} \hspace{5mm}

\caption{Tracking a rotating deformable surface. {\bf Top row:} Without rotation
handling, tracking with the original GBDF descriptors eventually breaks down. {\bf Bottom row:} Our modified GBDF features can track the whole sequence with up to $360^o$ of rotation.}
\label{fig:rotation_handling}
\vspace{-0.4cm}
\end{figure*}

We also demonstrate that the proposed rotation handling technique described in Section \ref{features} that overcomes the rotation sensitivity of the GBDF descriptors can successfully track a rotating deformable object.
\comment{We ran our tracking algorithm with and without rotation handling on the well-textured paper dataset with images rotated 5 degree every frame. This rotation in addition to original surface deformations introduce large frame-to-frame motions.}
Fig.~\ref{fig:rotation_handling} shows that without rotation handling, the original GBDF descriptors can only track up to 50 degrees of rotation, while the proposed rotation handling technique can track the whole sequence including a full 360 degrees of rotation.

\subsection{Sparsely Textured Surfaces}

\begin{figure} 
\centering
  \includegraphics[height=1.6cm]{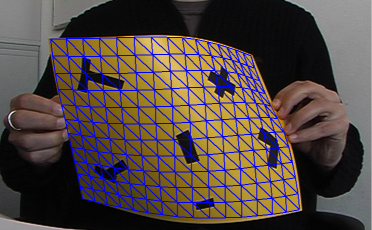} 
  \includegraphics[height=1.6cm]{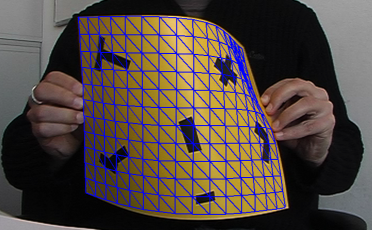} 
  \includegraphics[height=1.6cm]{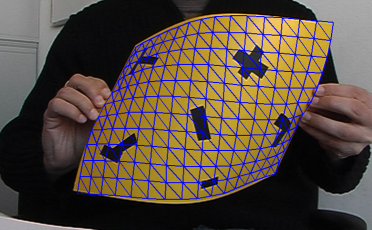}    \\
  \includegraphics[height=1.6cm]{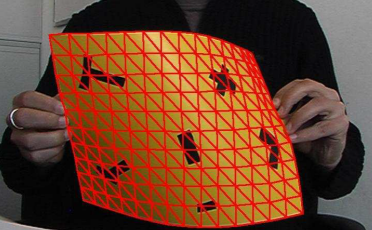}
  \includegraphics[height=1.6cm]{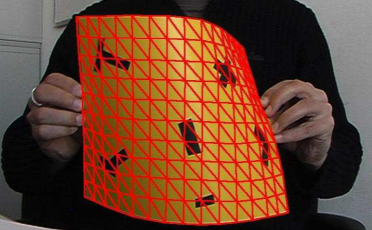}
  \includegraphics[height=1.6cm]{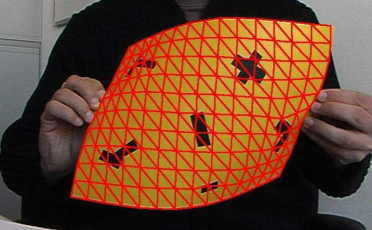}
  \caption{Sample reconstructions from the \cite{Salzmann08a} dataset.  While no ground truth is available in 3D, our results (top row) are qualitatively observed to be very accurate; the best published results on this dataset  are \cite{Salzmann08a} (bottom row), which has to extract the image edges explicitly, and involves learning, while our method does not. We do not have access to a reference image where the surface is in its planar rest shape, as our mesh assumes, causing some misalignment at the surface boundary.} 
  \label{fig:Salz_video_frames}
  \vspace{-2mm}
\end{figure}

To understand the performance of the methods in a realistic, sparsely textured setting, a different dataset is required.  A less textured paper dataset exists, as published in \cite{Salzmann08a}, but no ground truth information is available for this dataset in 3D, and so it is not suitable for numerical comparisons.  Nevertheless, for qualitative comparison purposes, we ran the proposed framework on this dataset, and our reconstructions align very well to the image information.  Example frames are provided in Fig.~\ref{fig:Salz_video_frames}, and the entire video is provided as supplementary material. The best known published results on this dataset are found in \cite{Salzmann12a}, which uses an algorithm that requires training data in addition to explicitly delineating the edges of the surface.  Our proposed framework is seen to perform as well as this previous method, qualitatively, while requiring no learning.

\begin{figure}
\vspace{-3mm}
\centering 
{\subfigure[GBDF+Oc]
{\includegraphics[height=1.8cm, width=2.3cm]{fig/white_paper/our_method_f226} 
\hspace{.1cm}}}
{\subfigure[NCC]
{\includegraphics[height=1.8cm, width=2.3cm]{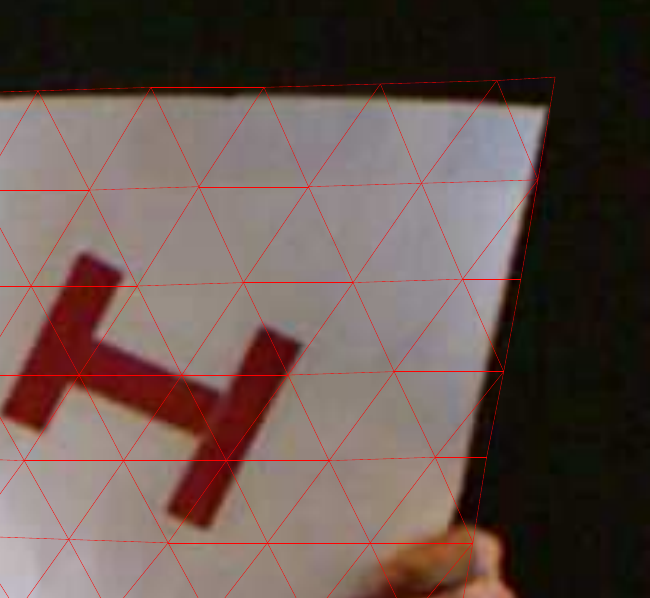}
\hspace{.1cm}}}
{\subfigure[MI]
{\includegraphics[height=1.8cm, width=2.3cm]{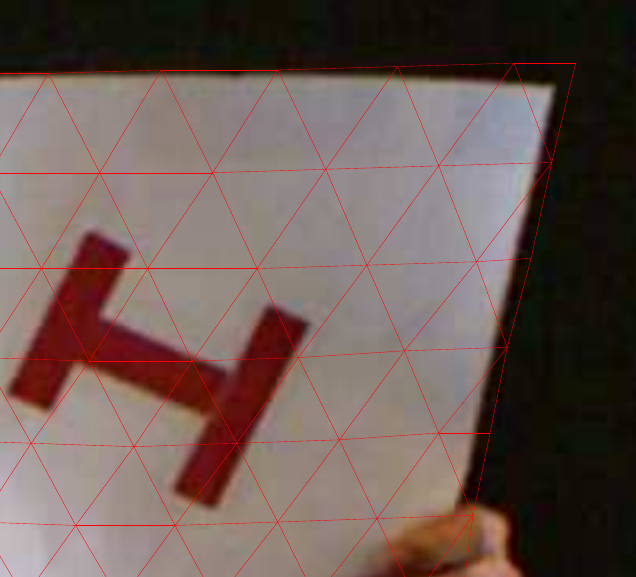}
\hspace{.1cm}}} \\ \vspace{-0.2cm}
{\subfigure[GBDF alone]
{\includegraphics[height=1.8cm, width=2.3cm]{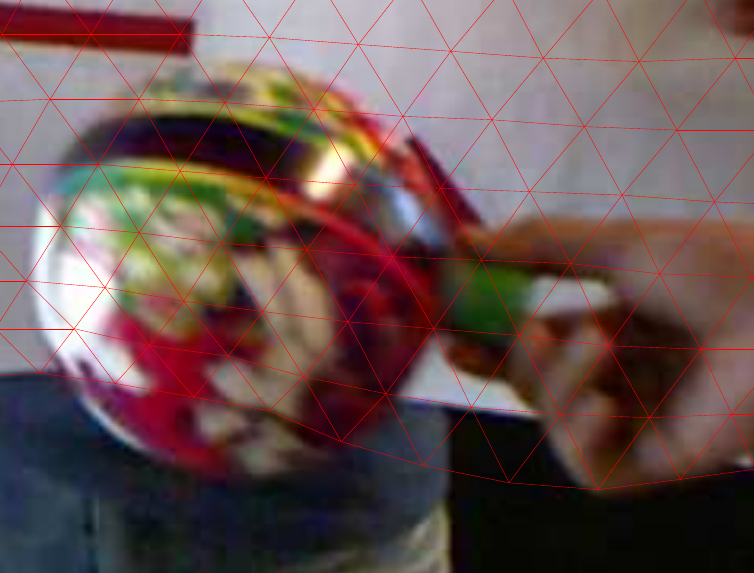}
\hspace{.1cm}}}
{\subfigure[Tukey] 
{\includegraphics[height=1.8cm, width=2.3cm]{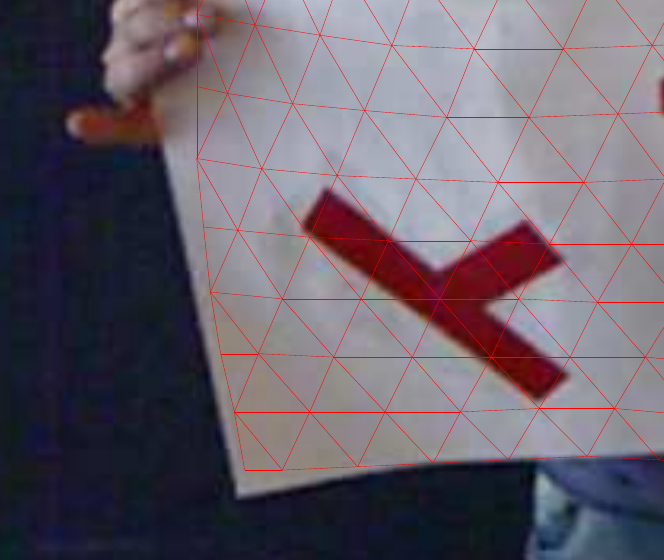}
\hspace{.1cm}}}
{\subfigure[Huber] 
{\includegraphics[height=1.8cm, width=2.3cm]{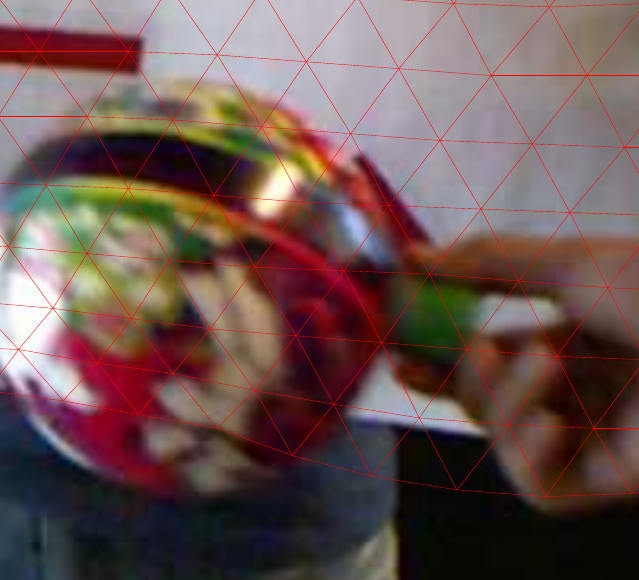}
\hspace{.1cm}}}
\caption{Reconstruction results on the same single frame. The proposed framework can track the whole sequence accurately, while other methods are seen to have trouble handling occlusions on top of the sparsely textured surface.}
\label{fig:white_paper_failure}
\vspace{-0.5cm}
\end{figure}

In order to be able to perform more meaningful numerical comparisons, we constructed a new dataset along with ground truth in 3D using a Kinect sensor, example images are provided in Fig.~\ref{fig:white_paper_failure}. \remove{This new dataset, along with ground truth, will be made available for public use. The} This new sparsely textured paper dataset contains various deformations and large lighting changes along with occlusions.

\begin{figure}
\centering
  \includegraphics[width=0.48\textwidth,height=3cm]{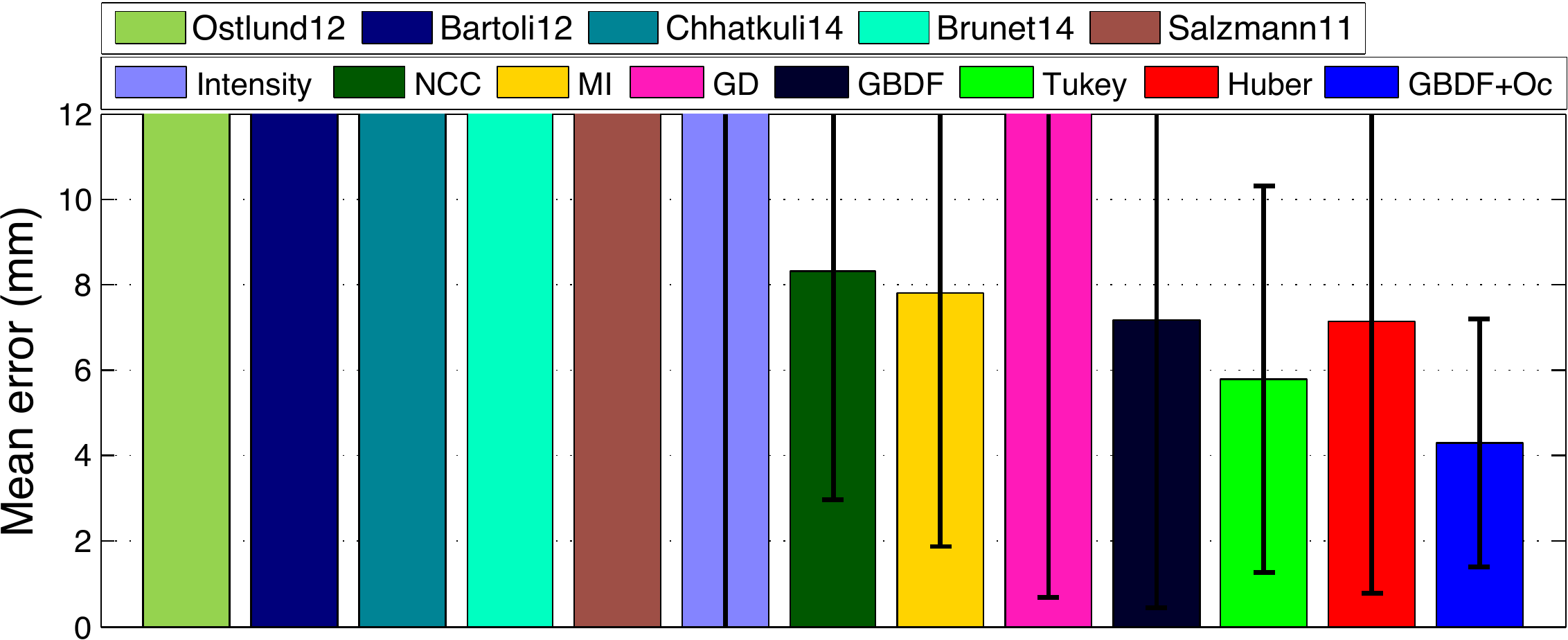}
  \caption{Reconstruction results on the sparsely textured paper dataset. Feature-based methods fail to reconstruct plausible surfaces, as indicated by the out-of-range error bars on the left.} 
  \label{fig:white_paper_errors}
  \vspace{-0.5cm}
\end{figure}
 
Quantitative results using the new dataset are presented in Fig.~\ref{fig:white_paper_errors}.
Feature-based methods that fail to reconstruct plausible surface shapes are indicated by high error bars that exceed y-axis range. Fig.~\ref{fig:white_paper_failure} provides a representative reconstruction on a single frame.  \ac{NCC} and \ac{MI} can track the surface fairly accurately, however they fail to capture fine details at the surface boundaries and hence the recovered depths in 3D are not very precise.  Without occlusion handling, dense matching with gradient-based descriptors often fails near occlusions.  The M-estimators are inconsistent near occlusions.  However, the proposed framework is seen to be able to accurately track the surface throughout the entire sequence.

It is interesting to note that while the occlusions are cleanly delineated in the relevancy score over a textured surface, they are much less obviously visible in the relevancy score over a sparsely textured surface.  This is expected, because well-textured un-occluded regions have consistently high correlation values with the template, and so it is only the occluded regions that are assigned low relevancy scores.  However, image regions of little texture have low and noisy correlation with a template, so occluded regions of similarly low correlation are assigned similarly low relevancy scores, and an occluded region is not as obviously distinct from the low textured regions in the relevancy score map.  This is one of the strengths of the proposed framework, because only the truly meaningful image regions are allowed to strongly influence the image energy cost.

\subsection{Applications}

\begin{figure} [b]
\vspace{-0.3cm}
\includegraphics[height=1.52cm]{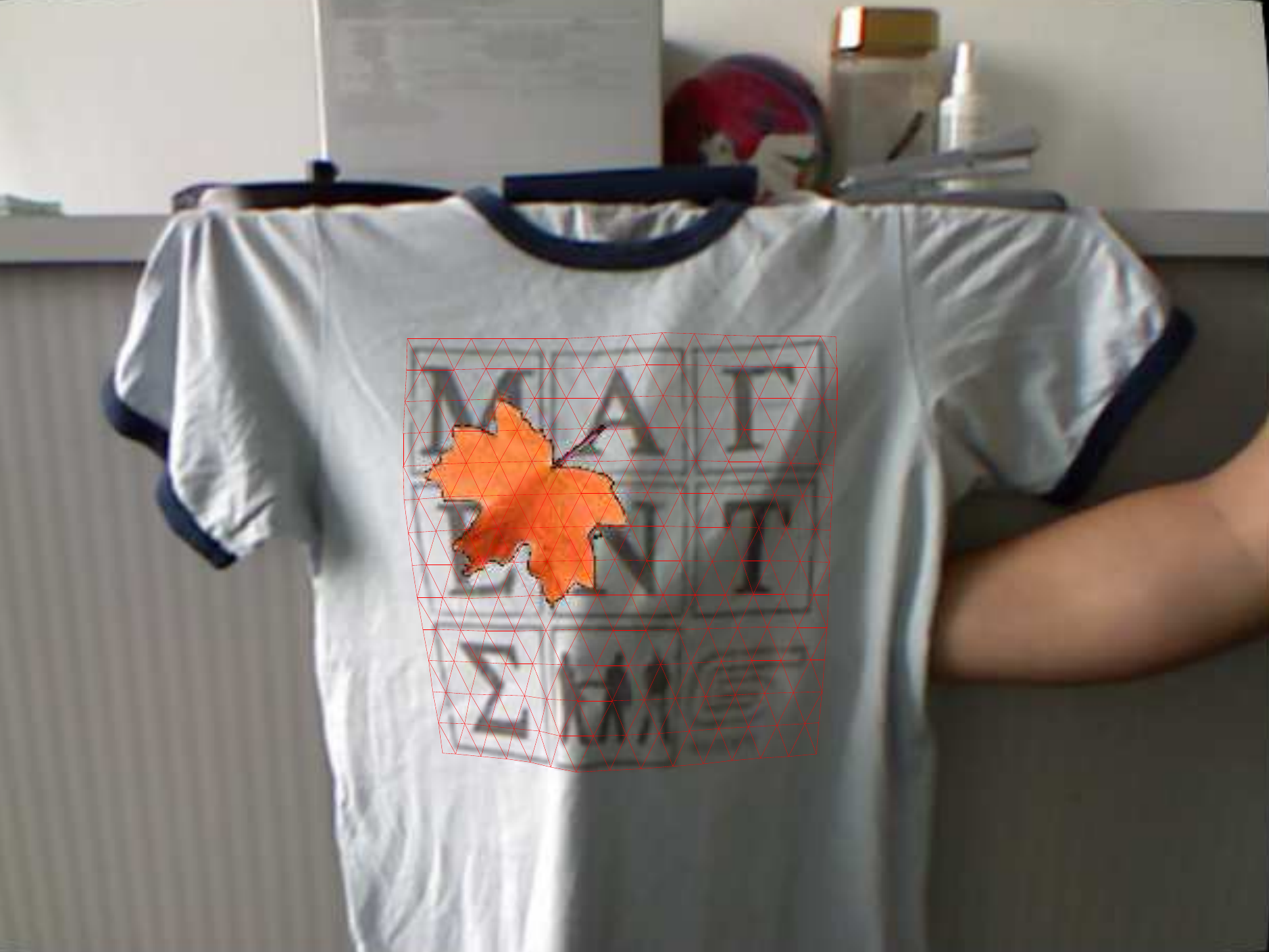} 
\includegraphics[height=1.52cm]{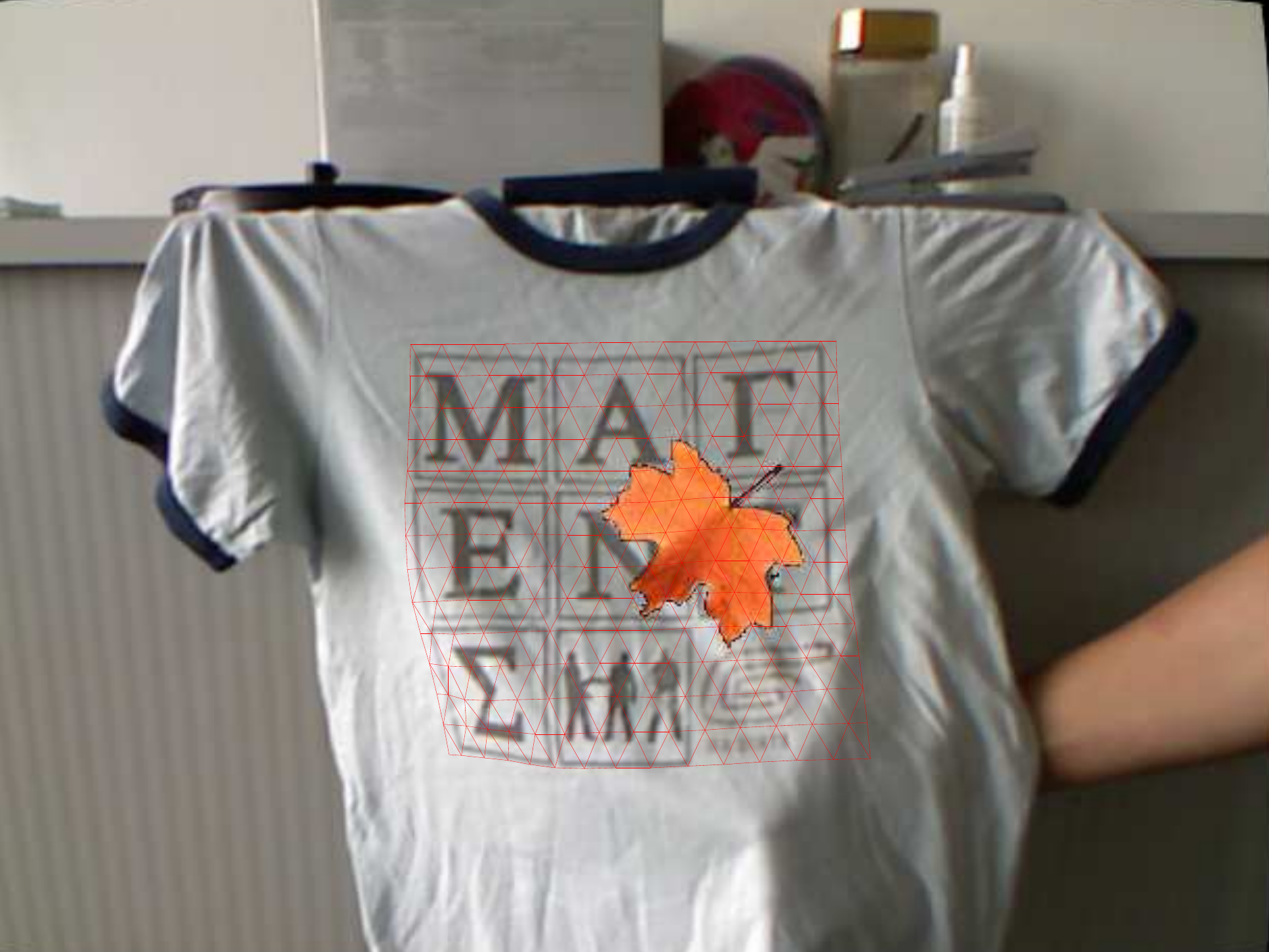} 
\includegraphics[height=1.52cm]{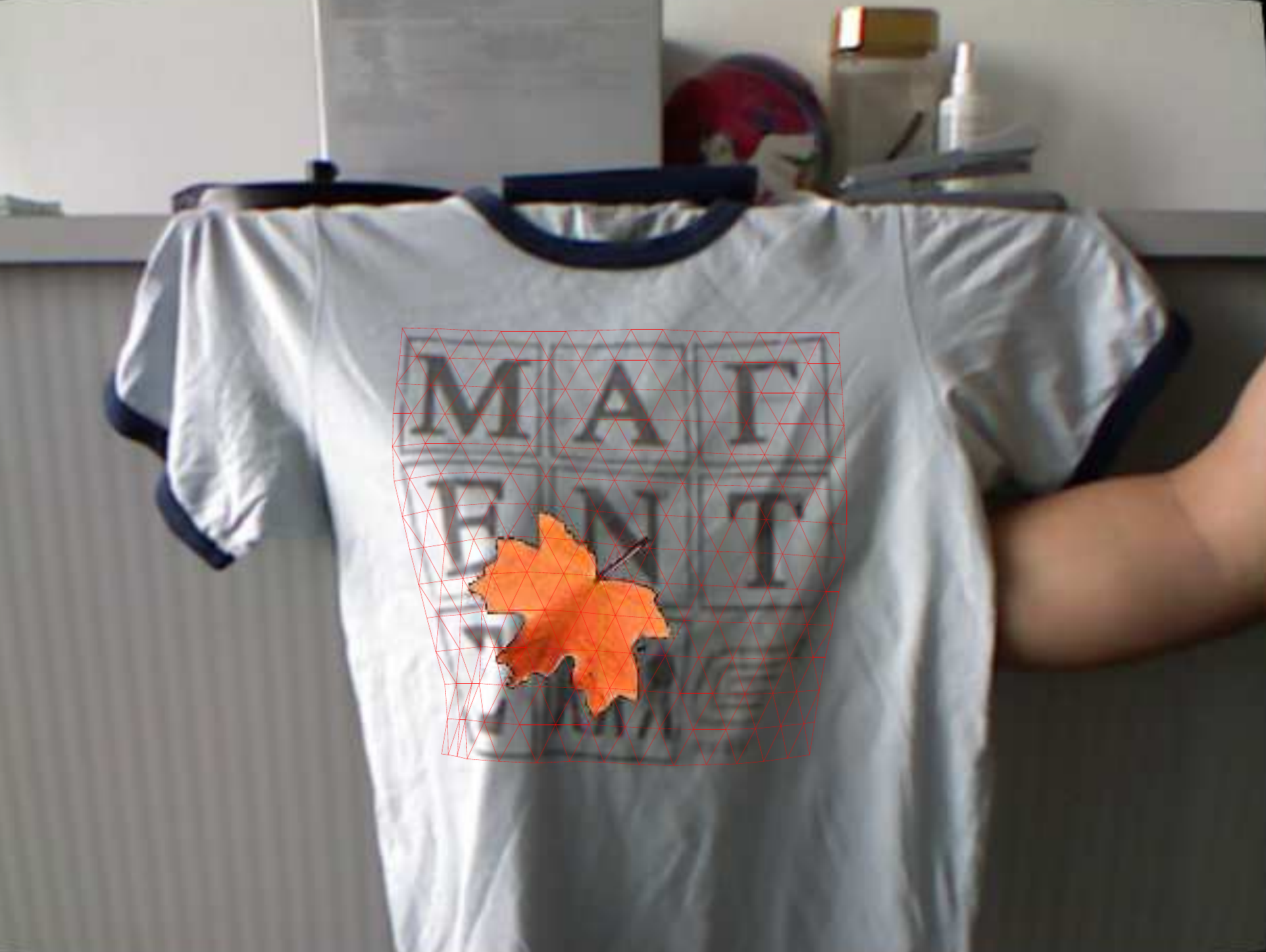}
\includegraphics[height=1.52cm]{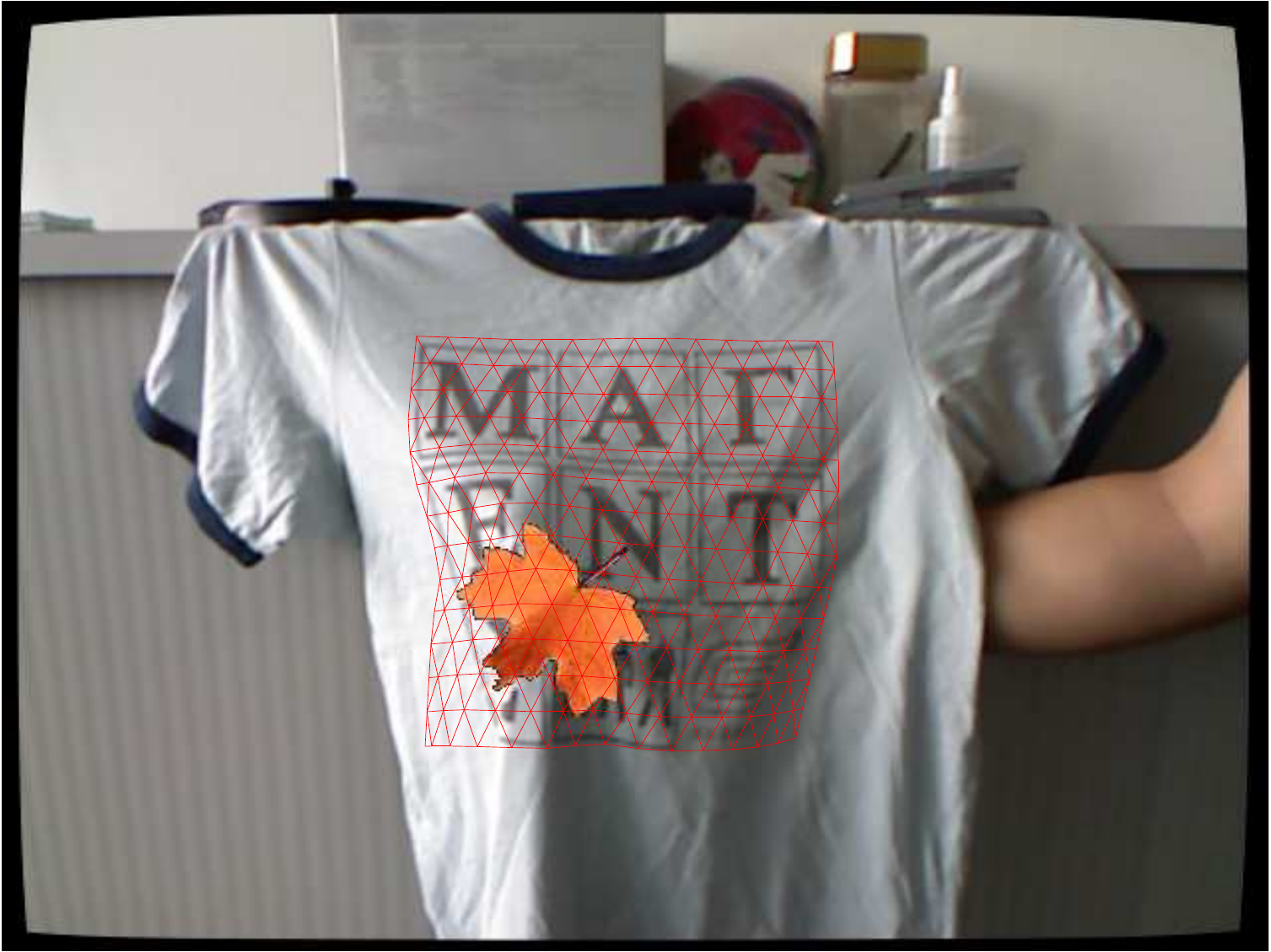}
\caption{Our representative reconstructions on the T-shirt dataset with artificial occlusions added. Rightmost: a tracking failure case when occlusions appear at areas with large deformations.}
\label{fig:tshirt_dataset}
\vspace{-0.2cm} 
\end{figure}

\begin{figure}
\vspace{-0.25cm} 
\centering 
\includegraphics[width=0.48\textwidth,height=3cm]{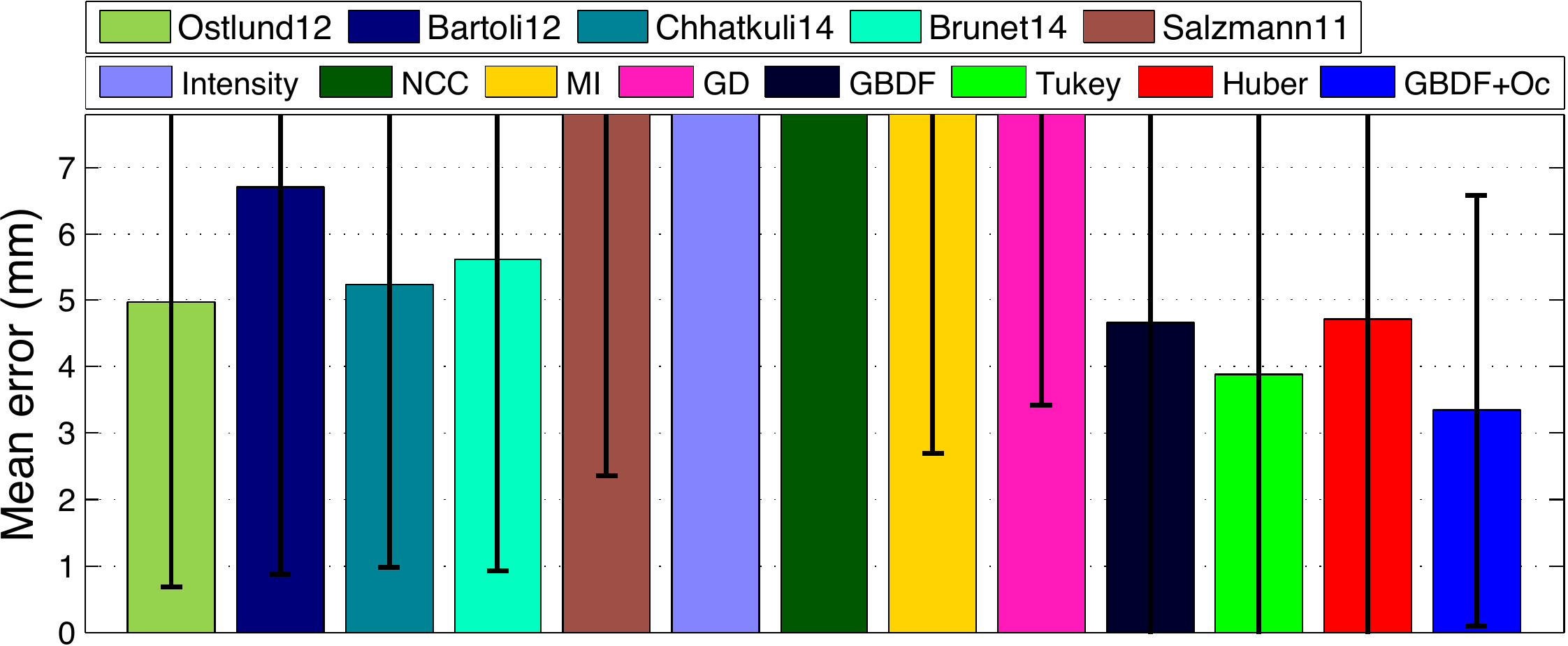}
\caption{Reconstruction results on the T-shirt dataset.}
\label{fig:tshirt_errors} 
\vspace{-0.25cm} 
\end{figure}

We demonstrate the robustness of our method in a variety of real-world applications.
%
%
%
First, we provide results on a cloth surface undergoing a different type of
deformation than studied in the paper datasets.  We created a new dataset along
with ground truth in 3D using a Kinect sensor, as before, to which artificial
occlusions were added, example images and our reconstructions are shown in
Fig.~\ref{fig:tshirt_dataset}. Quantitative results are presented in
Fig.~\ref{fig:tshirt_errors}.

\begin{figure}
\centering
  \includegraphics[height=3.4cm]{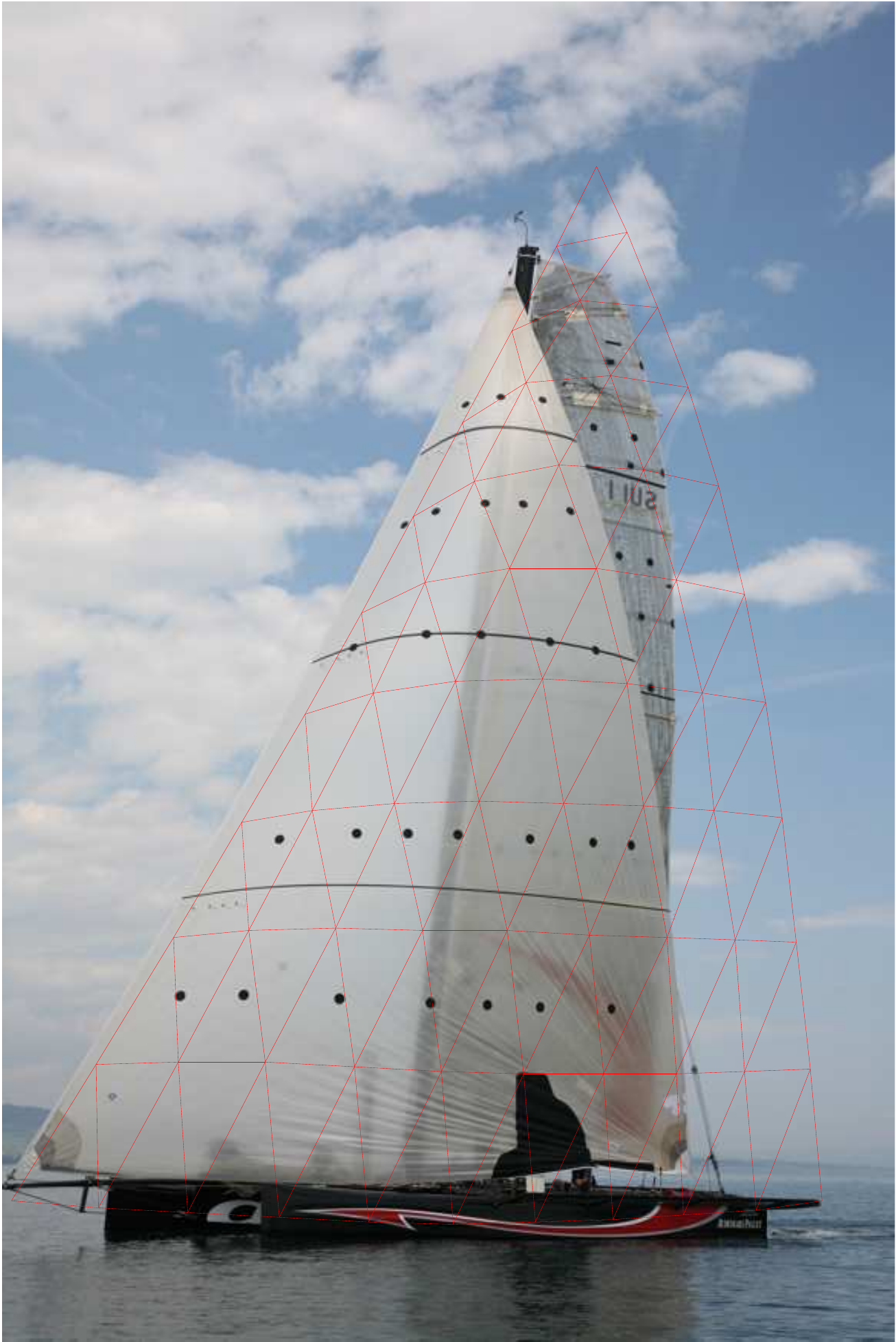}  \hspace{4mm}
  \includegraphics[height=3.4cm]{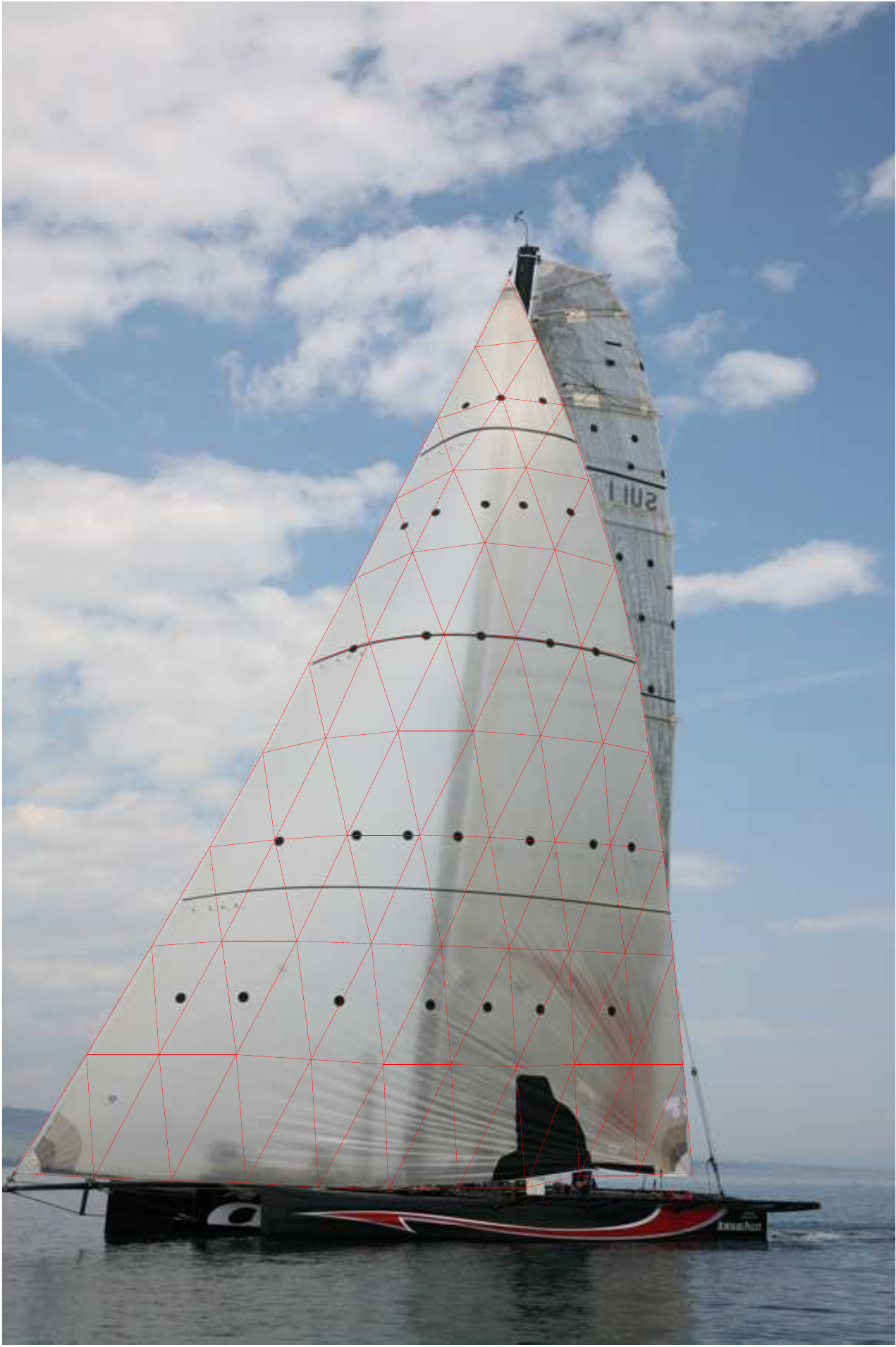} \hspace{4mm}
  \includegraphics[height=3.4cm]{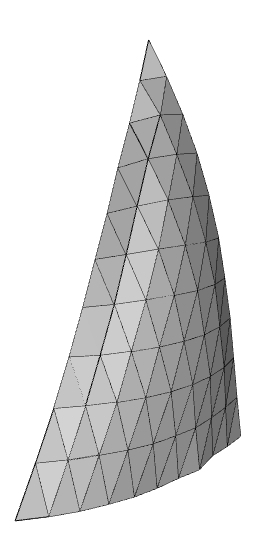}
  \caption{Image registration and surface reconstruction on the sparsely textured sail surface. From left to right: input image and initialization; final registration and reconstruction; the sail shape seen from a different viewpoint.}
  \label{fig:sail}
  \vspace{-2mm}
\end{figure}

The strength of our approach is demonstrated on a sparsely-textured sail surface
with a few dot markers, shown in Fig.~\ref{fig:sail}. Thanks to the large basin
of convergence of our algorithm, we can simply initialize the registration from
a very rough initial estimate without having first to establish correspondences.
Our algorithm naturally exploits line features, which feature point-based
methods usually do not.

\begin{figure} 
\centering
  \includegraphics[height=1.9cm]{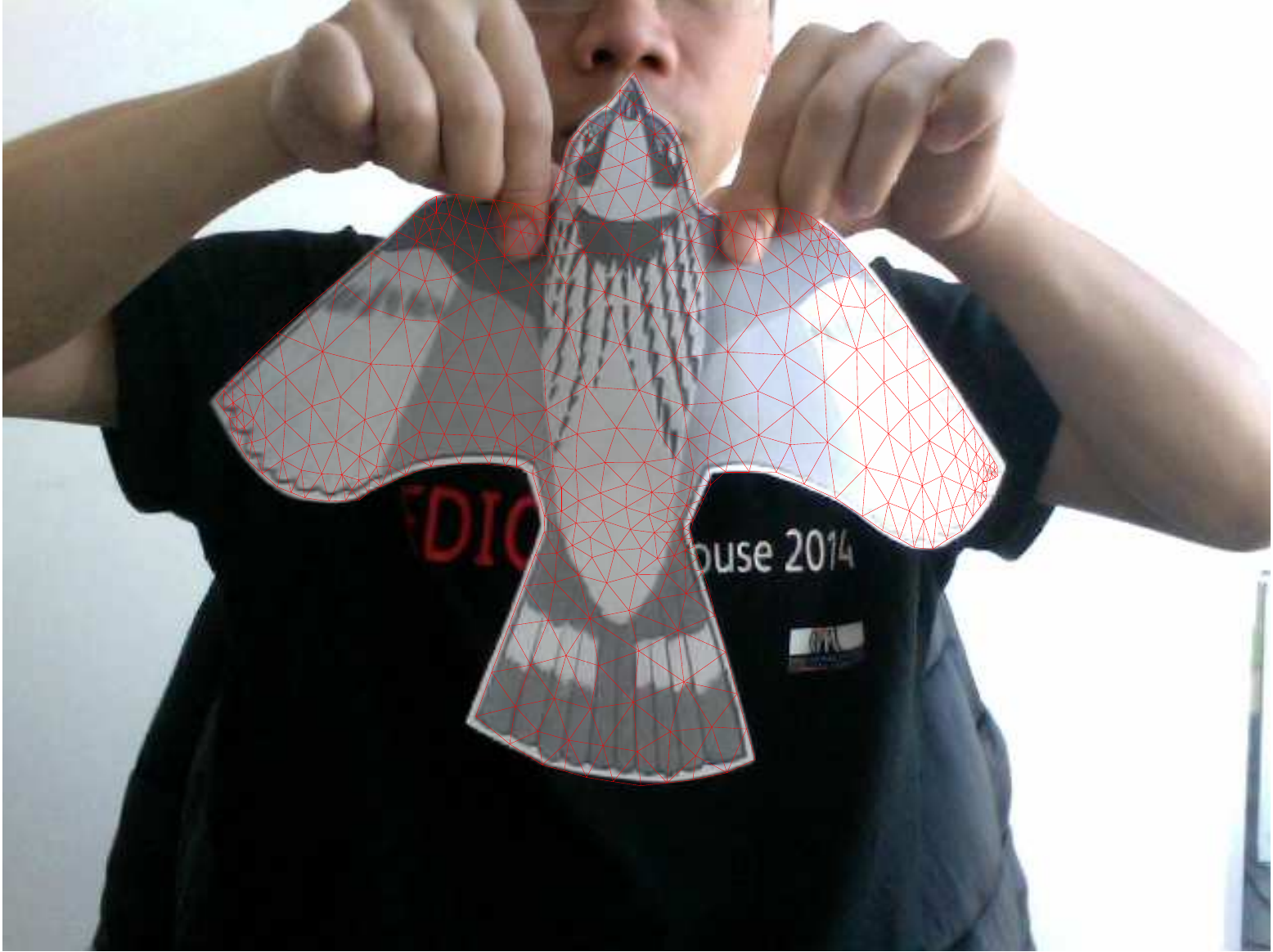} 
  \includegraphics[height=1.9cm]{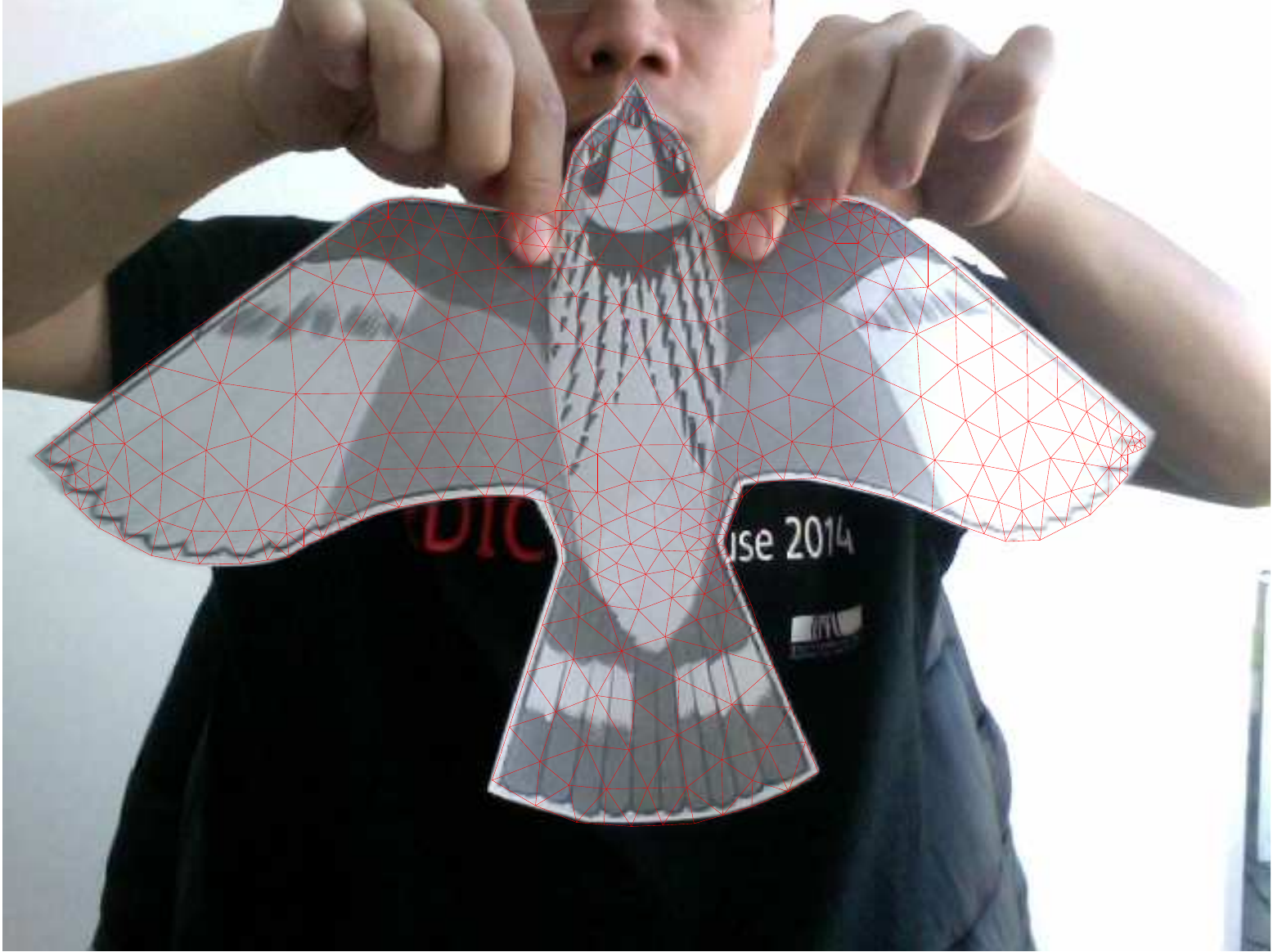} 
  \includegraphics[height=1.9cm]{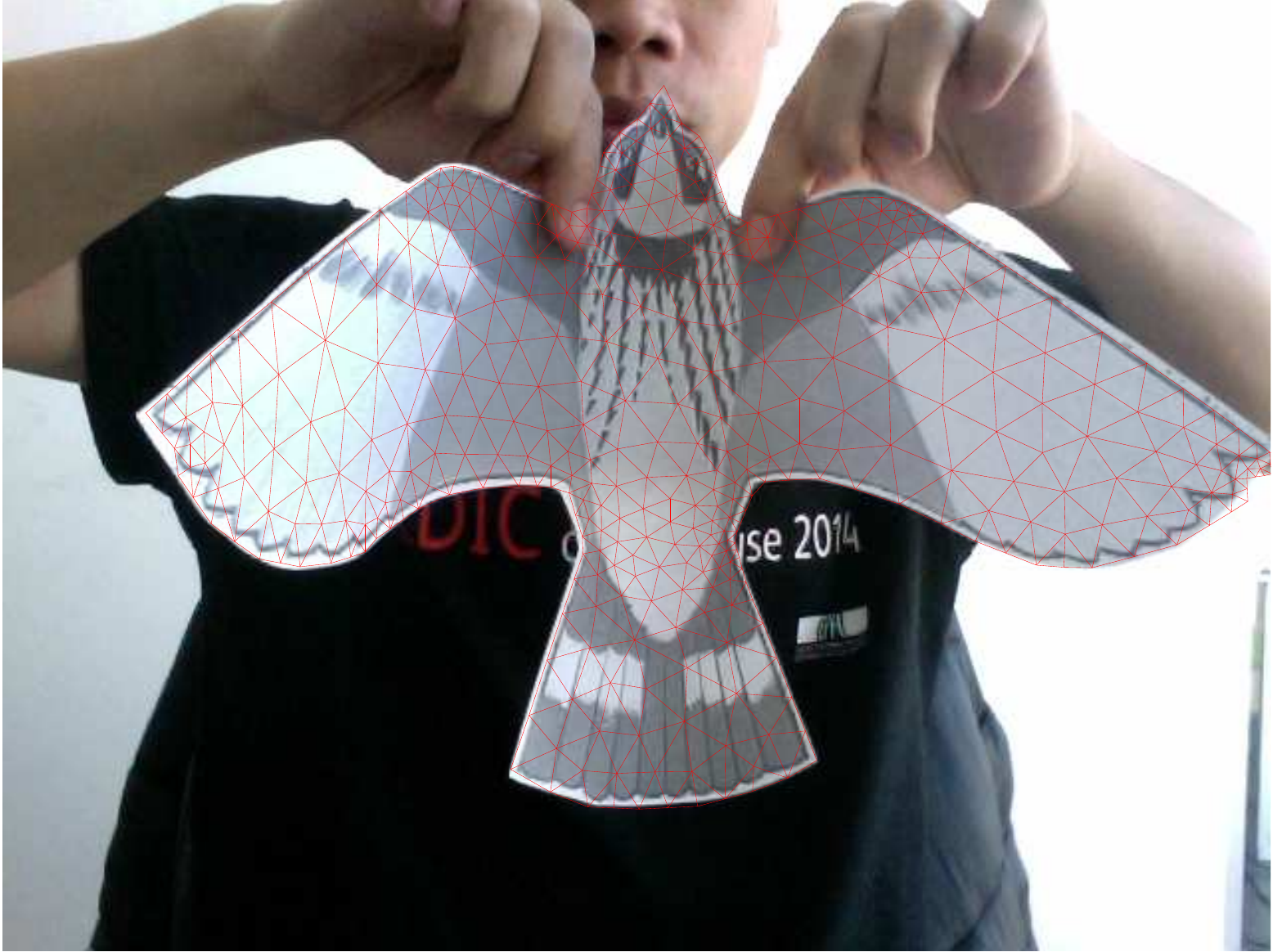} \\ \vspace{0.1cm}
  \includegraphics[height=3.2cm]{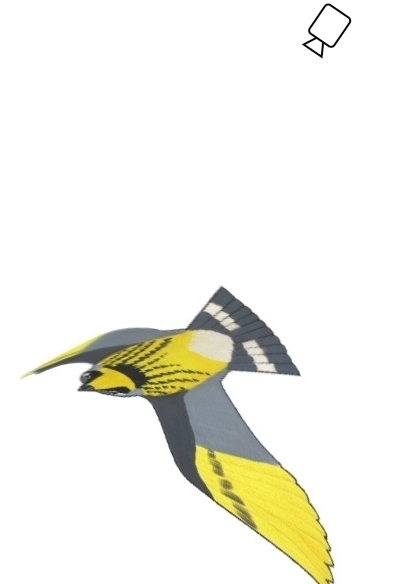} \hfill
  \includegraphics[height=3.2cm]{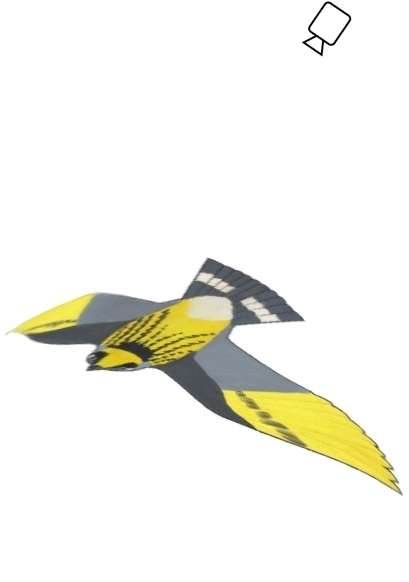} \hfill
  \includegraphics[height=3.2cm]{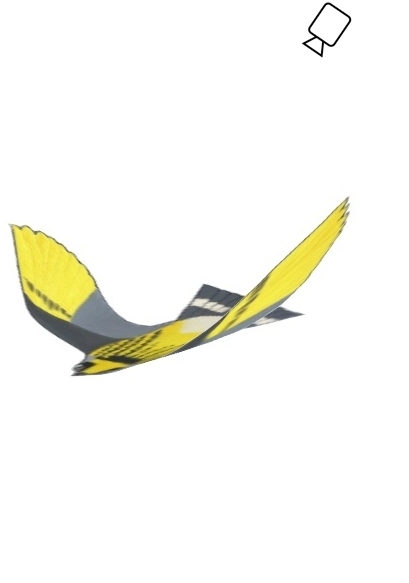}    
  \caption{Surface reconstruction of an animation capture from a monocular camera stream.} 
  \label{fig:bird}
  \vspace{-4mm}
\end{figure}

Fig.~\ref{fig:bird} depicts another application of our method for animation
capture from a monocular camera stream. In this setting, we capture the
animations of a bird whose animations can be transferred to another character.
The video of captured animations is provided in the supplementary material.

\section{Conclusion}

We have presented a framework for tracking both well textured and sparsely textured deforming surfaces in videos in the presence of occlusions.  Our framework computes a relevancy score for each pixel, which is then used to weight the influence of the image information from that pixel in the image energy cost function.  The presented method favorably compares to standard cost functions used for handling occlusion, such as Mutual Information and M-estimators.


\clearpage
{\small
\bibliographystyle{ieee}
\bibliography{short,vision,biomed,graphics,learning,optim,robotics}
}  

\end{document}